\documentclass[11pt]{article}

\usepackage[final]{acl}

\usepackage{times}
\usepackage{latexsym}

\usepackage[T1]{fontenc}
\usepackage[utf8]{inputenc}

\usepackage{microtype}
\usepackage{inconsolata}

\usepackage{graphicx}
\usepackage{booktabs}
\usepackage{xcolor}
\usepackage{amsmath}
\usepackage{amssymb}
\usepackage{makecell}
\usepackage{amsmath,amsfonts,bm,mathtools,amssymb}
\usepackage{array}
\usepackage{subcaption}
\usepackage{rotating}
\usepackage{multirow}

\newcommand{\hidden}[1]{}

\title{Are Linguistic and Arithmetic Circuits Dissociated in LLMs?}

\author{
  \textbf{Riku Kisako\textsuperscript{1}},
  \textbf{Tatsuki Kuribayashi\textsuperscript{2,3}},
  \textbf{Ryohei Sasano\textsuperscript{1}} \\
  \textsuperscript{1}Nagoya University, Japan \quad
  \textsuperscript{2}MBZUAI, UAE \quad
  \textsuperscript{3}Tohoku University, Japan \\
  \texttt{\{kisako.riku.n3@s.mail, sasano@i\}.nagoya-u.ac.jp} \\
  \texttt{tatsuki.kuribayashi@mbzuai.ac.ae}
}
\begin{document}
\maketitle

\begin{abstract}
  Large language models (LLMs) exhibit both knowledge \textit{of} language, rendered by, for example, grammatical judgment ability, and abilities to perform tasks \textit{through} language, such as math-word problem solving.
Motivated by neurolinguistic studies of their neural dissociation in human brains, in this study, we investigate the overlap of their circuits in large language models (LLMs).
Specifically, we examine the overlap and causal effects between the circuits extracted for linguistic and arithmetic tasks, using controlled materials and circuit discovery methods from the mechanistic interpretability field.
Our experiments with five LLMs show that circuit similarity is mostly explained by task domains (linguistic vs. arithmetic), rather than by their input and output vocabularies or by the syntactic similarities between natural-language sentences and arithmetic expressions.
This finding suggests that LLMs may process linguistic and arithmetic information using largely separate neural circuits, highlighting an ability-based modular organization of circuits within these models.\footnote{
Code is available at \url{https://github.com/Riku-0801/linguistic_arithmetic_circuit.git}
}

\end{abstract}

\section{Introduction}
\label{sec:intro}

\begin{figure}[t]
  \centering
  \includegraphics[width=0.48\textwidth]{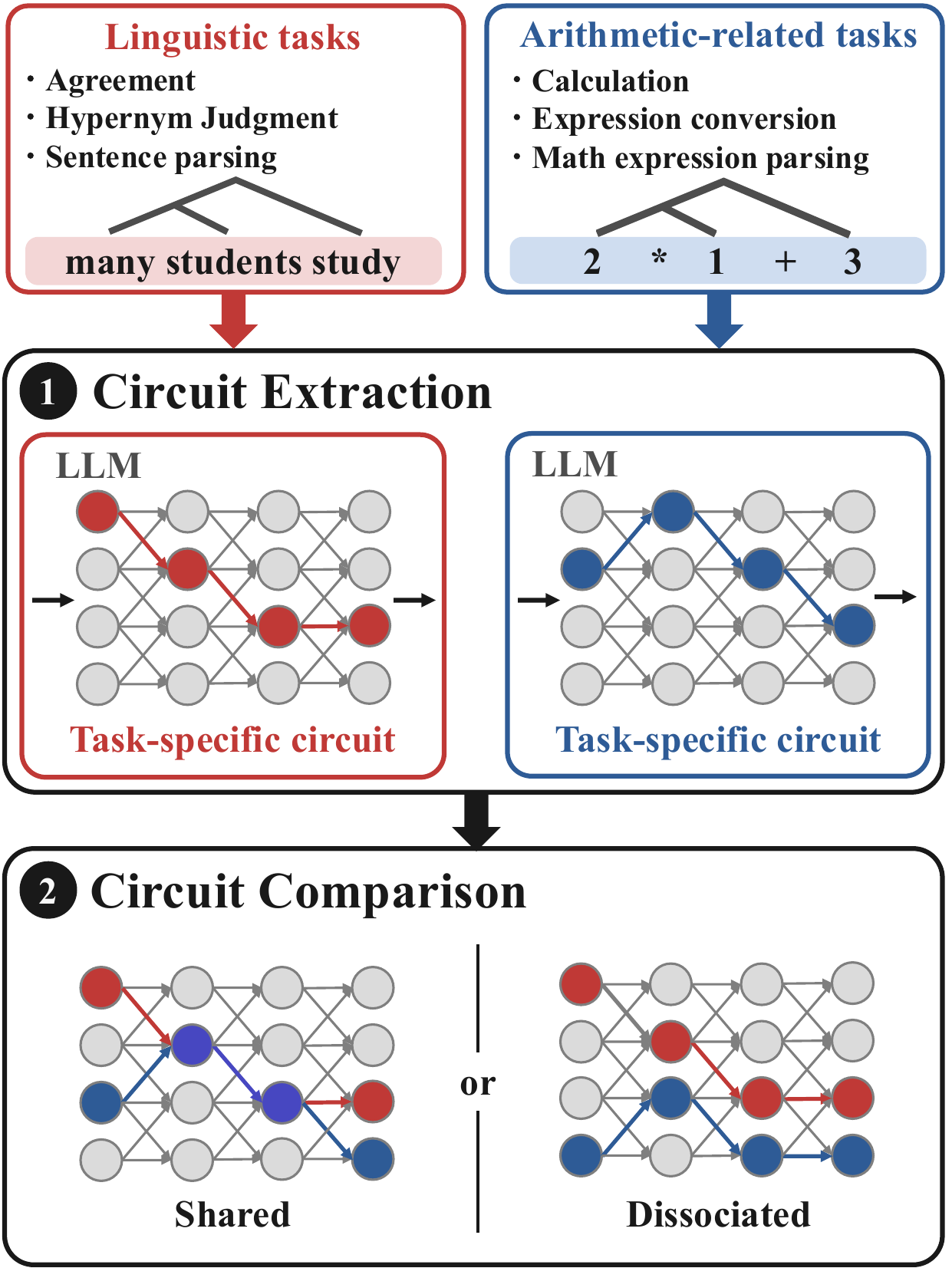}
  \caption{Overview of our circuit analysis. We extract a task-specific circuit for each linguistic and arithmetic-related task and compare the extracted circuits to examine whether they are shared or dissociated across the two domains.}
    \label{fig:fig1}
\end{figure}

Large language models (LLMs) generally exhibit both \emph{formal linguistic competence}, involving grammatical and lexical knowledge of language itself, and \emph{functional linguistic competence}, involving broader abilities performed through language~\cite{MAHOWALD2024517}.
One scientific question is how these different types of abilities are implemented and interact within LLMs.
This is motivated by neurolinguistic human studies that seek to answer whether the brain areas for language and complex thought are overlapping, and recent studies have suggested their dissociation~\cite{Fedorenko2010,Fedorenko2011,varley2005agrammatic}, while some studies report certain interactions~\cite{Nakai2018}.
Compared to human experiments, LLM studies enable fine-grained internal circuit analysis and more controlled experiments, and hopefully give insights into what-if questions, for example, whether a human-like, potentially modular organization of formal and functional linguistic competence naturally emerges from statistical learning of language without explicit architectural constraints for neural LLMs.
\citet{hanna-etal-2026-formal}, for example, have investigated the overlap between formal and functional task circuits within LMs, using a broad range of tasks. 

In this study, we complement such existing works with a more fine-grained, controlled investigation on a specific task. 
In particular, the stability of task circuits across different surface formats has been underexplored.
In addition, functional tasks typically involve multiple subtasks, and it is important to ablate the circuits for each component of the task.
We specifically employ \emph{arithmetic} tasks to represent functional linguistic competence, which are also a typical testbed in human studies~\cite{Fedorenko2011, Nakai2018}, to enable flexible control of vocabulary alternation, e.g., Arabic numeral vs. number word, and subtask isolation, e.g., with and without arithmetic operation on the same arithmetic expressions.
Motivated by the long-running debates of the shared structural, recursive nature of natural language and arithmetic expressions~\cite{Chomsky2006-qp, amalric2018cortical,Matsumoto2023-nl}, we also compare the parsing settings on natural language and arithmetic expressions with essentially the same syntactic constructions, exemplified by \texttt{Cat (deeply sleeps)} vs. \texttt{(Cute Cat) sleeps}; and \texttt{2 + (3 * 1)} vs. \texttt{(2 * 3) + 1} as a potentially maximal interactions between these two domains.
We test whether LMs reuse shared circuits to process these similar syntactic structures across different domains of language and arithmetic, motivated by human studies showing cross-domain interactions~\cite{Nakai2018}.

In our experiments, we extract an LLM circuit (i.e., a subnetwork) for each task using Edge Attribution Patching with Integrated Gradients (EAP-IG) and compare their structural overlap and cross-task faithfulness (Figure~\ref{fig:fig1}). 
Our experiments demonstrate the following characteristics of the circuits for arithmetic tasks and their interaction with linguistic ones:
\begin{itemize}
    \item Similar arithmetic circuits are robustly obtained among different surface forms (e.g., Arabic-numeral and number-word inputs), indicating that input/output vocabulary alone does not determine the arithmetic circuit.
    \item Different circuits are obtained for different task instructions (e.g., just spell-out or do calculation) on the same arithmetic expressions, further supporting that arithmetic circuits are mainly determined by the required ability, not by input alone.
    \item Arithmetic circuits do not causally interact with linguistic circuits, even for the essentially similar task of parsing natural language sentences with similar syntactic complexity to the used arithmetic expressions, supporting that linguistic and arithmetic circuits are largely dissociated within LMs.
\end{itemize}

\section{Related Work}
\subsection{Mechanistic Interpretability and Circuit Analysis}
The mechanistic interpretability field aims to explain model behavior in terms of internal components and the computations they implement. 
The community has found that model components responsible for specific task abilities can be localized as a subnetwork, i.e., circuit, of the model~\cite{elhage2021mathematical,olsson2022induction}.
Prior work has explored localization in the LLM parameter space, e.g., through learning binary masks over LLM weights~\cite{chen-etal-2024-learnable}, or intervening in the activation space~\cite{dai-etal-2022-knowledge}.
More recently, \emph{circuit analysis} methods have been widely used to extract task-relevant sub-architecture in transformers~\cite{heimersheim2024useinterpretactivationpatching,syed-etal-2024-attribution}.
Causal intervention techniques, such as activation patching, has recently been employed to identify components relevant to a target behavior.
The obtained circuits are typically analyzed through the lens of their architectural overlap, and the importance of causal metrics such as cross-task faithfulness has also been emphasized~\cite{hanna2024have}, which this study also relies on. 

\subsection{Arithmetic and Formal Linguistic Competence in Humans}
Neurolinguistic research has localized language-specific brain regions, supporting the view that language processing is not reducible to domain-general cognition~\cite{Fedorenko2010,Fedorenko2011,Shain2020-lq,Hu2023-ca}.
Still, researchers have studied the extent to which these language networks are indeed limited to natural language, with arithmetic as a particularly debatable case. 
The field of linguistics has highlighted the fundamental similarity between language and arithmetic/algebraic expressions, especially given their recursive nature~\cite{Chomsky2006-qp,Matsumoto2023-nl}.
This raises the question of whether the language network is also involved in arithmetic processing.

There is accumulating evidence supporting that natural language and arithmetic processing are \textit{dissociated}.
\citet{varley2005agrammatic}, for example, showed that patients with severe agrammatic aphasia could nevertheless perform arithmetic operations involving hierarchical and recursive structures, suggesting that an intact language-processing network is not necessary for arithmetic computation in the mature cognitive system.
\citet{friedrich2009mathematical} examined the syntactic processing of formulae in first-order mathematical logic and identified a network primarily involving intraparietal and prefrontal regions, with only selective involvement of Broca's area. \citet{maruyama2012cortical} further showed that the processing of nested arithmetic expressions predominantly recruits bilateral posterior occipitotemporal and parietal regions rather than the left-lateralized language network. 
\citet{amalric2018cortical} identified a bilateral network centered on the intraparietal sulci and inferior temporal regions that is consistently recruited by mathematical reflection, whereas classical language and semantic regions show little involvement. 
Subsequent fMRI experiments with professional mathematicians showed that this mathematical network responds to mathematical statements across domains and independently of several possible confounds, including task difficulty, logical operators, and whether the relevant knowledge is retrieved or computed~\cite{amalric2019distinct}.
More broadly, \citet{Fedorenko2024-dd} argues that language is mainly for communication, not for complex thought, such as arithmetic operations.

At the same time, several studies have also demonstrated cross-domain structural interaction between natural language and arithmetic processing~\cite{Nakai2018,Zhou2018-uv,Nakai2023-oe}.
For example, \citet{Nakai2018} demonstrates a syntactic priming effect between natural language and arithmetic expressions.
Note that there are general distinctions among (i) arithmetical computation, such as \texttt{(2+3)*4}, (ii) abstracted algebraic computation, such as \texttt{(a+b)*c}, and (iii) mathematical problem solving, such as math word problems.
This study mainly focuses on the dissociation between natural language and arithmetical computation.

\subsection{Arithmetic and Formal Linguistic Competence in LLMs}
What about the circuit dissociation between language and arithmetic processing?
A growing body of work studies whether LLMs represent distinct circuits for different domains. 
Recent work has examined whether formal and functional linguistic mechanisms~\cite{MAHOWALD2024517} are dissociated in LLMs, providing evidence that linguistic form and broader task-oriented processing need not rely on the same internal representations~\cite{hanna-etal-2026-formal}. 
\citet{kisako2025representationaldissociationlanguagearithmetic} study representational geometry using linear classifiers and cluster separability, whereas we compare edge circuits under interventions. 
\citet{alkhamissi-etal-2025-llm} analyzes language-selective units within LLM, motivated by neuroscientific studies of humans, and the involvement of LLM language networks in other tasks, such as math problems. 
Yet, their analysis does not precisely test the robustness of the circuits against various input/output formats and sub-task decomposition.

More broadly, the natural language processing field has investigated both the sources and mechanisms of mathematical ability in LLMs. 
Benchmark studies have shown that strong performance on math word problems can mask fragile or shortcut-based behavior~\cite{li-etal-2024-gsm}. 
Other work has suggested that arithmetic- or code-heavy pretraining data can substantially affect downstream capability profiles~\cite{aryabumi2025to,petty-2025-code,razeghi2024backtracking}. 
At the mechanistic level, recent analysis has argued that arithmetic behavior in LLMs is often supported by heuristic mechanisms rather than robust algorithmic procedures~\cite{nikankin2025arithmetic}.

\section{Circuit Analysis in LLMs}
\label{sec:circuit}

Circuit analysis for transformer LLMs~\cite{NIPS2017_3f5ee243} treats a model as a directed computational graph from inputs to outputs through intermediate components, and aims to identify a subgraph (a \emph{circuit}) that is sufficient for a target behavior.
Hereafter, \textit{nodes} correspond to modules (e.g., attention heads or MLP blocks), and \textit{edges} correspond to information flow between nodes~\cite{wang2023interpretability}.

\subsection{Method}
We use EAP-IG (Edge Attribution Patching with Integrated Gradients)~\cite{hanna2024have}.
EAP-IG computes an \emph{indirect effect} score for each directed edge $e_{i \rightarrow j} \in \mathcal{E}$.
To compute the score, for each task, we construct paired inputs $(x, \tilde{x})$, where $x$ and $\tilde{x}$ are called clean and corrupted input, respectively.\footnote{
Our analysis focuses on the behavior of $ x$ at the last time step $t$, but for brevity, let us omit $t$ from the notation, i.e., $x$ typically means $x_t \in x$.} 
The corresponding representation at node $i$ is denoted as $\bm h_i(x) \in \mathbb{R}^n$, and the token representation as $\bm x \in \mathbb{R}^m$, respectively.
Given a differentiable metric $m(\cdot): \mathbb{R}^m\to \mathbb{R}$ (the output is typically a logit), the gradient at the receiving node $j$, given an input token representation $\bm x$, is denoted as $\nabla_{\mathbf{h}_{j}}m\left(\bm x\right) \in \mathbb{R}^n$.
The sender activation difference is $\Delta \bm h_i=\bm h_i(x)-\bm h_i(\tilde{x}) \in \mathbb{R}^n$.
By discretizing the input interpolation path with $T$ steps, we approximate the integrated gradient score as:
\begin{align}
s_{i \rightarrow j} 
&= \Delta \bm h_i^{\top}\int_{\alpha=0}^1\nabla_{\bm h_j}m\!\left(\tilde{\bm x} + \alpha(\bm x - \tilde{\bm x}) \right)\,\mathrm{d}\alpha \;\;\mathrm{,} \\
&\approx \frac{\Delta \bm h_i^{\top}}{T}\sum_{k=0}^{T-1}
\nabla_{\bm h_j}m\!\left(\tilde{\bm x} + \tfrac{k}{T}(\bm x - \tilde{\bm x}) \right) \;\;\mathrm{,}
\end{align}
where we average gradients at interpolated inputs and take their inner product with the sender activation difference.
The metric $m$ is, for example, in the subject--verb agreement (SVA) task, the logit difference between singular and plural verbs: $m(\bm x) = \mathrm{logit}(\mathrm{are}|\bm x) - \mathrm{logit}(\mathrm{is}|\bm x)$.
The corrupted input $\Tilde{\bm x}$ is designed so that the correct output in $\bm x$ becomes incorrect; for example, for a clean input \textit{``The keys on the cabinet [are],''} the corrupted input can be \textit{``The key on the cabinet [is].''}
We extract a circuit $E$ by selecting edges with high attribution scores.
To test whether our conclusions depend on how the circuit boundary is defined, we report results under three circuit settings:
(i) top $2.5\%$ edges, (ii) top $5.0\%$ edges, and (iii) an adaptive circuit defined as the smallest edge set whose self-faithfulness exceeds $0.85$.
We then analyze the overlap between extracted circuits $E_A$ and $E_B$ for tasks $A$ and $B$.

\begin{table*}[t]
  \centering
  \footnotesize
  \setlength{\tabcolsep}{0.15cm}
  \begin{tabular}{@{}p{4cm} p{5cm} p{1.8cm} c l c@{}}
    \toprule
    \textbf{Task} &
    \textbf{Input (Prefix)} &
    \textbf{Answer} &
    \textbf{Input} &
    \textbf{Output} &
    \textbf{Arith.} \\
\midrule
    Arithmetic calculation. (\textsc{Math}) &
    The answer to \texttt{2 * 1 + 3 =} is &
    5 &
    Expr. &
    Arabic numeral &
    \checkmark \\
\cmidrule(r){2-3} \cmidrule(r){4-5} \cmidrule(lr){6-6}
    Spelled-out arithmetic calculation (\textsc{Math-s}) &
    The answer to \texttt{two times one plus three equals} is &
    five &
    Lang. &
    Number word &
    \checkmark \\
    \cmidrule(r){2-3} \cmidrule(r){4-5} \cmidrule(lr){6-6}
    Math Word Problem \newline (\textsc{GSM-Plus}) &
    Janet's ducks lay 16 eggs per day... The answer is &
    18 &
    Lang. &
    Arabic numeral &
    \checkmark \\
    \midrule
    Arithmetic expression spelling-out (\textsc{Spell-m}) &
    The spelled-out form of \texttt{2 * 1 + 3 =} is &
    two \textcolor{gray}{times one plus three} &
    Expr. &
    Number word &
     \\
\cmidrule(r){2-3} \cmidrule(r){4-5} \cmidrule(lr){6-6}
    Inverse of \textsc{Spell-m} \newline (\textsc{Spell-mi}) &
    The numerical formula for \texttt{two times one plus three} is &
    2 \textcolor{gray}{* 1 + 3} &
    Lang. &
    Arabic numeral &
     \\
\cmidrule(r){2-3} \cmidrule(r){4-5} \cmidrule(lr){6-6}
    Arithmetic expression parsing \newline (\textsc{Math-Parse}) &
    The first term to calculate in 6 + 2 * 5 is &
    2 \textcolor{gray}{* 5} &
    Expr. &
    Arabic numeral & (\textcolor{gray}{\checkmark})
    \\
\midrule
    Sentence parsing \newline (\textsc{Ling-Parse}) &
    The head word of the lowest constituent of the sentence 'many students study' is &
    students &
    Lang. &
    Word &
     \\
\cmidrule(r){2-3} \cmidrule(r){4-5} \cmidrule(lr){6-6}
    Subject--Verb Agreement \newline (\textsc{Ling-SV}) &
    The keys on the cabinet &
    are &
    Lang. &
    Word &
     \\
\cmidrule(r){2-3} \cmidrule(r){4-5} \cmidrule(lr){6-6}
    Gendered Pronoun Agreement \newline (\textsc{Ling-GA}) &
    Maria said that &
    she &
    Lang. &
    Word &
     \\
\cmidrule(r){2-3} \cmidrule(r){4-5} \cmidrule(lr){6-6}
    Hypernym Judgment \newline (\textsc{Ling-HJ}) &
    Roses, and other &
    flower &
    Lang. &
    Word &
     \\

    \bottomrule
  \end{tabular}
  \caption{Task examples in our experiments. Horizontal blocks indicate four high-level categories based on task domain and required competence.
 Expr. and Lang. denote symbolic expressions and natural language, respectively. 
  For generation tasks, we evaluate only the first generated token;
  subsequent tokens are shown in gray.
  Arithmetic checkmark denotes whether the task involves arithmetic calculation, and \textsc{Math-Parse} has a gray one, as it does not involve calculation but should be a necessary sub-task for arithmetic.
  }
  \label{tab:datalist}
\end{table*}

\subsection{Metrics}
\label{subsec:metrix}

\paragraph{Structural overlap (IoU).}
To quantify overlap between the \emph{edge sets} of two task circuits $E_A$ and $E_B$, we use Intersection over Union (IoU):
\begin{equation}
\mathrm{IoU}(A,B)=\frac{|E_{A}\cap E_{B}|}{|E_{A}\cup E_{B}|}\,\,\mathrm{.}
\end{equation}

\paragraph{Functional overlap (Faithfulness).}
Even when structural overlap is small, circuit polysemy may allow a circuit extracted for one task to support another task.
To quantify how well the model behavior for a reference task $A$ is reproduced by
the circuit extracted from a different task $B$, we use the Faithfulness score
\cite{hanna2024have}.

Let $(\bm x^{A}$, $\tilde{\bm x}^{A}) \in \mathcal{D}^A$ be the paired clean and corrupted input representation for task $A$, and $\mathcal{D}^A$ is the dataset of task $A$.
We denote $E_B \subseteq \mathcal{E}$ as the set of edges in the circuit extracted for task $B$.
Henceforth, we simplify the notation of edge as $e \in \mathcal{E}$, and let $a_e(\bm x)$ be the edge activation/message
used in EAP-style edge patching when running the model on input $\bm x$.
We define a \emph{cross-task circuit application} to task $A$ using $E_B$ as follows:
we run the model on the \emph{clean} input $\bm x^{A}$ while \emph{patching} the activations
on all edges outside $E_B$ with those from the corrupted run on $\tilde{\bm x}^{A}$,
keeping the edges in $E_B$ unpatched (i.e., clean).
Concretely, we construct mixed edge activations as follows:
\begin{equation}
a^{(A \leftarrow B)}_e(\bm x^A) \;=\;
\begin{cases}
a_e(\bm x^{A}) & \text{if } e \in E_B\;\;\mathrm{,}\\
a_e(\tilde{\bm x}^{A}) & \text{if } e \notin E_B\;\;\mathrm{.}
\end{cases}
\label{eq:mixed_edge_acts}
\end{equation}
Setting this activation $a^{(A \leftarrow B)}_e(\bm x^A)$ to all the edges $e\in \mathcal{E}$ of the model, we input $\bm x^A$ and compute the metric $m^{(A \leftarrow B)}(\bm x^A)$ for an input of task $A$.
The transferred object is the edge mask $E_B$, not activations from task $B$: both clean and corrupted runs use task-$A$ inputs. A self-circuit need not reproduce the full-model score exactly.
The scores $m(\bm x^A)$ and $m(\tilde{\bm x}^A)$ denote the metric on the clean and corrupted inputs of task $A$ without any model intervention, respectively.
The faithfulness score $F(A,B)$ is computed as follows:
\begin{equation}
\frac{1}{|\mathcal{D}^A|}\sum_{(\bm x^A, \tilde{\bm x}^A)\in \mathcal{D}^A}
\frac{m^{(A \leftarrow B)}(\bm x^A) - m(\tilde{\bm x}^A)}{m(\bm x^A) - m(\tilde{\bm x}^A)}\;\;\mathrm{,}
\label{eq:faithfulness}
\end{equation}
where $\mathcal{D}^A$ is the dataset of task $A$.

We call $F(A,B)$ \emph{raw faithfulness} when distinguishing it from the additional display normalization; it is already normalized by clean/corrupted baselines and is not a raw logit score.
For cross-task Faithfulness analyses using the top-$2.5\%$ and top-$5.0\%$ circuit definitions, we additionally normalize $F(A,B)$ by the target task's self-faithfulness $F(A,A)$, yielding $F(A,B)/F(A,A)$. 
Values below zero or above one indicate performance below the corrupted or above the clean baseline, respectively, when the clean--corrupted gap is positive; they are not probabilities.
Small baseline gaps, or small diagonal values in the heatmaps, can amplify scores. Thus, absolute metric values and unscaled faithfulness are needed to interpret apparent transfer.

Further details on circuit extraction, cross-task evaluation, and clustering settings are provided in Appendix~\ref{sec:appendix-exp-settings}.

\section{Experimental Setup}
\label{sec:exp}

\subsection{Tasks}
\label{subsec:task}
We introduce a set of controlled tasks to avoid confounding factors, e.g., simple vocabulary overlap.
Table~\ref{tab:datalist} summarizes the tasks adopted in our experiments.
For all tasks, we use 500 clean/corrupted input pairs. 
We extract a circuit for each task and compute cross-task circuit overlap (IoU) and faithfulness scores.
We briefly introduce the tasks below; Appendix~\ref{sec:tasks-detail} provides more details, including prompt templates and clean/corrupted examples.

\paragraph{Pure arithmetic tasks.}
The pure arithmetic tasks are \textsc{Math} and \textsc{Math-s}, where \textsc{Math-s} is written in number words, instead of Arabic numerals and arithmetic symbols.
Arithmetic expressions contain three terms with addition, subtraction, and multiplication, and answers are constrained to be natural numbers.

\paragraph{Pure linguistic tasks.}
At the other end of the spectrum, following~\citet{hanna-etal-2026-formal}, we adopt three tasks of \textsc{Ling-SV}, \textsc{Ling-GA}, and \textsc{Ling-HJ} as purely linguistic tasks.
They focus on Subject--Verb Agreement (SV), Gendered Pronoun Agreement (GA), and Hypernym Judgment (HJ) tasks, respectively, and models are expected to answer with a plausible single word given context. 
Subject--verb agreement probes grammar, gendered pronoun agreement also involves syntactic dependences between referential and gender cues, and hypernym judgment probes lexical and semantic knowledge.

Other tasks are a mixture of the arithmetic problem and its subproblems.

\paragraph{Math word problems.}
We experimentally include math word problems (\textsc{GSM-Plus})~\cite{li-etal-2024-gsm} written in natural language, which arguably require both arithmetic and general text comprehension.
We use GSM-Plus data, built from GSM8K~\cite{cobbe2021trainingverifierssolvemath}, a benchmark of grade-school math word problems, by applying controlled perturbations, such as numerical and operator transformations, to test the robustness of mathematical reasoning in LLMs.
In our setup, we construct clean/corrupted pairs using the original (seed) problems and their numerical substitution variants from GSM-Plus.

\paragraph{Arithmetic expression without calculation.}
To control for math-domain input without requiring calculation, we include the spelling-out task for math expressions \textsc{Spell-m} and its inverse \textsc{Spell-mi}.
The model is instructed to convert an expression such as \texttt{2*1+3} into its spelled-out English form (e.g., \textit{two times one plus three}) without performing computation.
\textsc{Spell-mi} is a task to convert a spelled-out expression into an Arabic-numeral formula.
As the input equation is the same as \textsc{Math} tasks, comparing their circuit clarifies whether the circuit is simply determined by input similarity.

\paragraph{Parsing mathematical and linguistic materials.}
To explore the potential cross-domain similarity between language and arithmetic processing, we introduce \textsc{Math-Parse} and \textsc{Ling-Parse} tasks.
\textsc{Math-Parse} asks for operator precedence without any calculation; for example, the answer for the question, \texttt{The first term to calculate in 6 + 2 * 5 is}, is \texttt{2 * 5}.
On the other hand,  \textsc{Ling-Parse} asks for constituent structure, specifically the \emph{head word} of the lowest constituent,  in short sentences.
Each pair shares a noun and a verb: ``many students study'' targets \emph{students}, whereas ``students often study'' targets \emph{study}. 
We evaluate the logit of the first prediction token in the main experiments, and we designed the problems so that the first token can distinguish the model's parsing decision, and the correct/incorrect options are plausible in the sense that they both appear in clean and corrupted sentences.
See Appendix~\ref{sec:sequence-evaluation} for sequence-level evaluation.

\begin{table*}[t]
  \centering
  \scriptsize
  \setlength{\tabcolsep}{3pt}
  \resizebox{\textwidth}{!}{%
  \begin{tabular}{@{}lrrrrrrrrrr@{}}
    \toprule
    Model
      & \textsc{Math}
      & \textsc{Math-s}
      & \textsc{GSM-Plus}
      & \textsc{Spell-m}
      & \textsc{Spell-mi}
      & \textsc{Math-Parse}
      & \textsc{Ling-Parse}
      & \textsc{Ling-SV}
      & \textsc{Ling-GA}
      & \textsc{Ling-HJ} \\
    \midrule
    Llama-3.1-8B & 96.0 & 85.2 & 11.4 & 100.0 & 100.0 & 99.8 & 94.8 & 99.8 & 92.6 & 98.8 \\
    Qwen3-8B & 90.8 & 77.2 & 36.8 & 100.0 & 100.0 & 100.0 & 90.0 & 100.0 & 86.2 & 97.6 \\
    Qwen3-1.7B & 82.2 & 45.4 & 23.0 & 99.6 & 100.0 & 100.0 & 84.0 & 43.2 & 79.4 & 29.6 \\
    Qwen3-4B & 92.0 & 60.0 & 36.8 & 99.8 & 100.0 & 100.0 & 93.0 & 86.0 & 79.6 & 78.6 \\
    Gemma-2-2B & 66.6 & 41.2 & 24.2 & 100.0 & 100.0 & 61.4 & 75.2 & 99.6 & 85.2 & 98.0 \\
    \bottomrule
  \end{tabular}%
  }
  \caption{Model accuracy on each task. Values are percentages. We report exact top-1 accuracy, except in cases where only the available answer-vs.-corrupted
pairwise rows are used in place of exact accuracy.}
  \label{tab:model-accuracy}
\end{table*}

\begin{figure}[t]
  \centering

  \begin{subfigure}[t]{\linewidth}
    \centering
    \includegraphics[width=\linewidth]{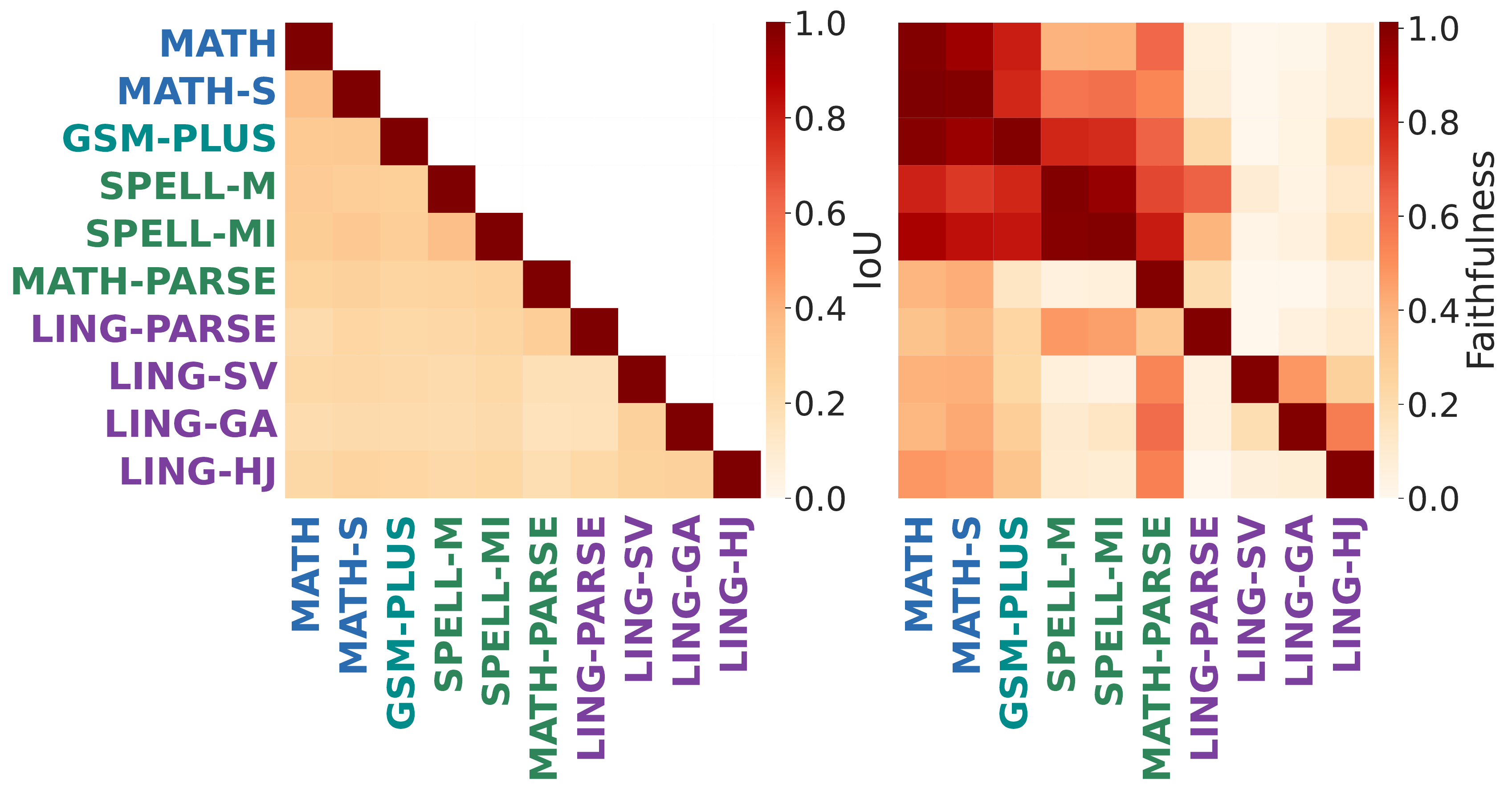}
    \caption{Qwen3-8B, top-2.5\% circuit.}
    \label{fig:qwen3-8-npmi-faithfulness}
  \end{subfigure}

  \vspace{1em}

  \begin{subfigure}[t]{\linewidth}
    \centering
    \includegraphics[width=\linewidth]{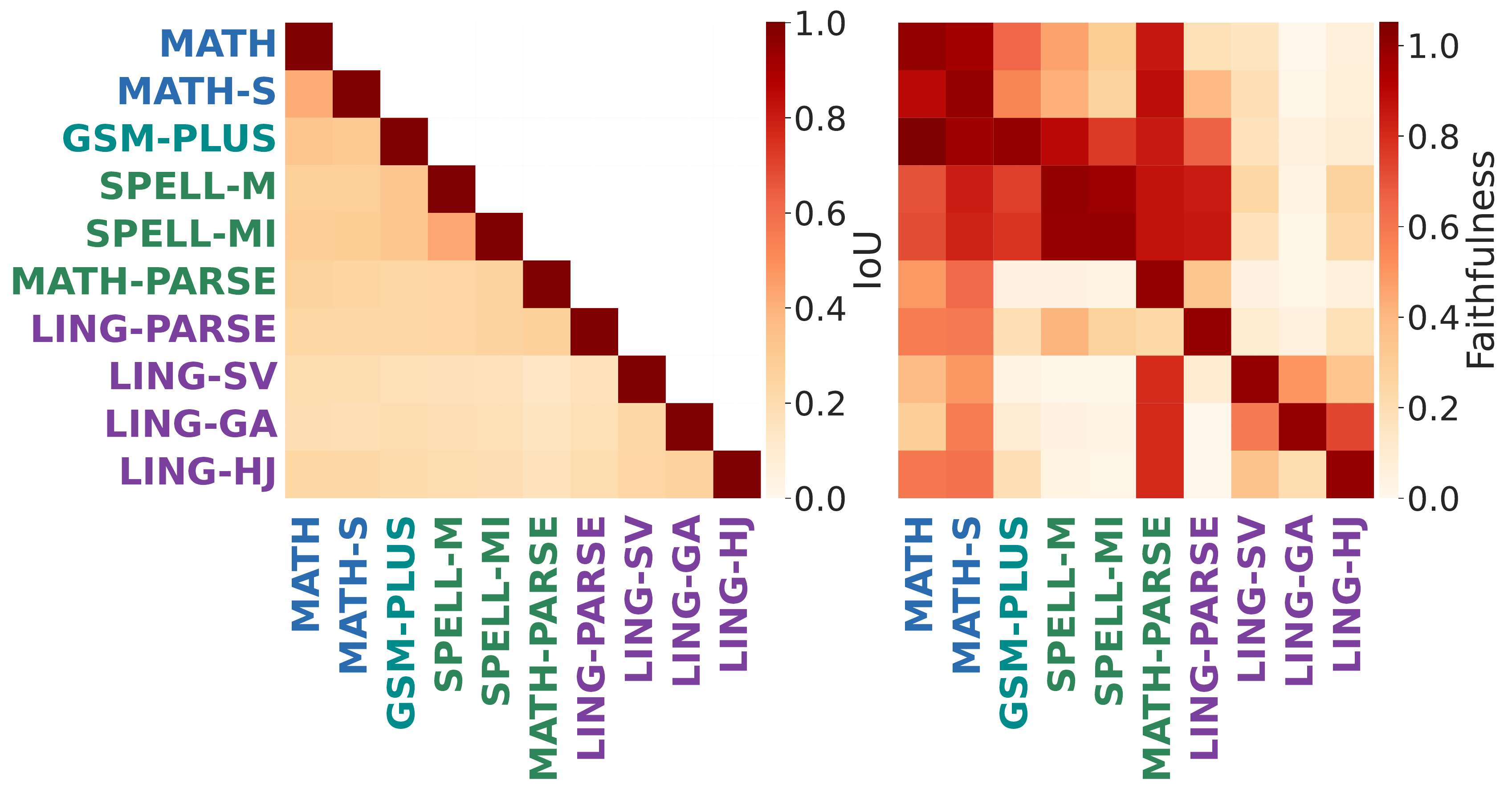}
    \caption{Llama-3.1-8B, top-2.5\% circuit.}
    \label{fig:llama-3.1-8b-npmi-faithfulness}
  \end{subfigure}
  \caption{Left: IoU (structural overlap). Right: cross-task Faithfulness (functional overlap). For Faithfulness, rows denote the tasks from which circuits are extracted, and columns denote the target tasks solved with those circuits. Scores are normalized by the corresponding diagonal self-faithfulness value.}
  \label{fig:npmi-faithfulness-all}
\end{figure}

\begin{figure}[t]
  \centering
  \includegraphics[width=\linewidth]{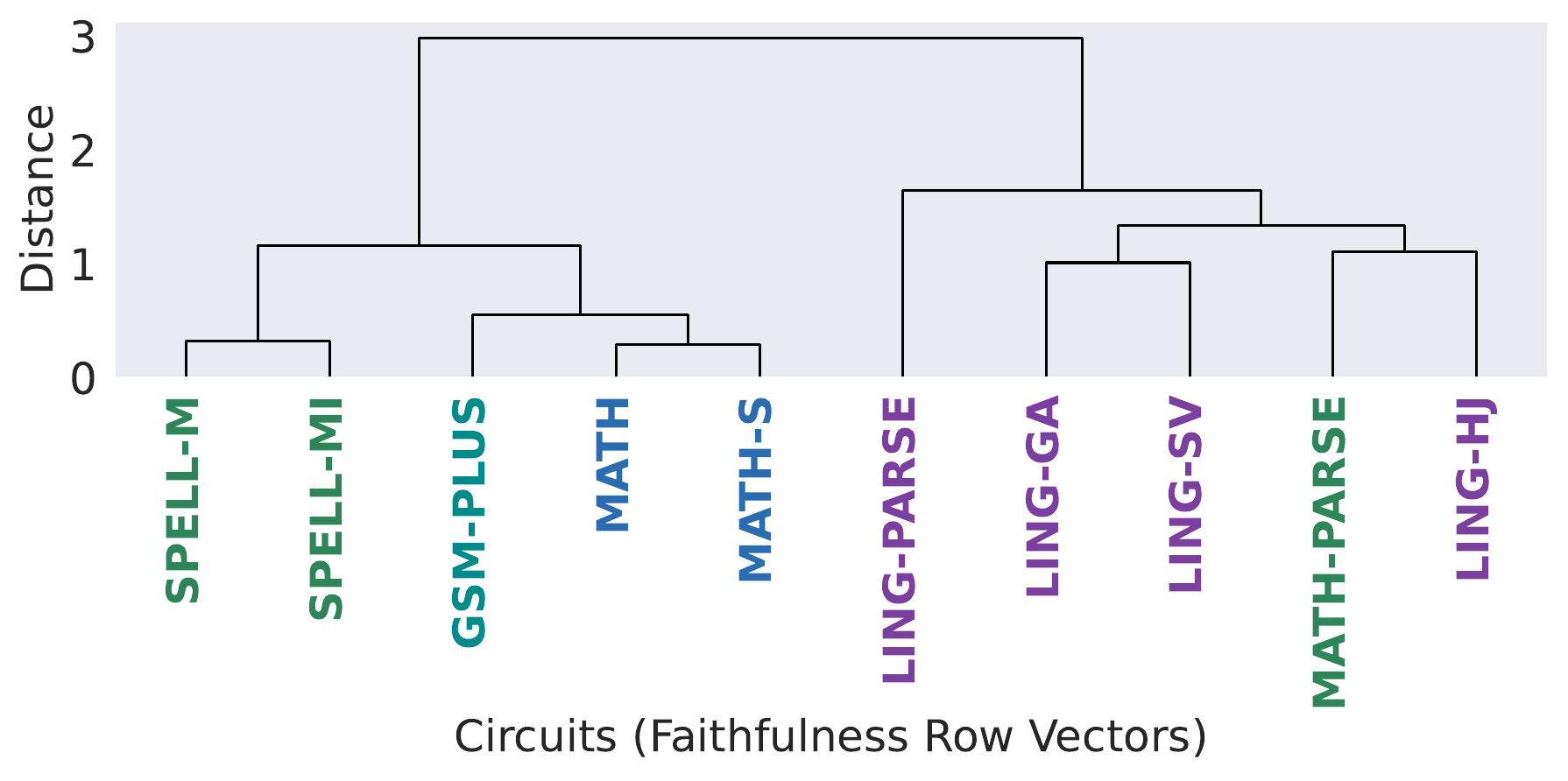}
  \includegraphics[width=\linewidth]{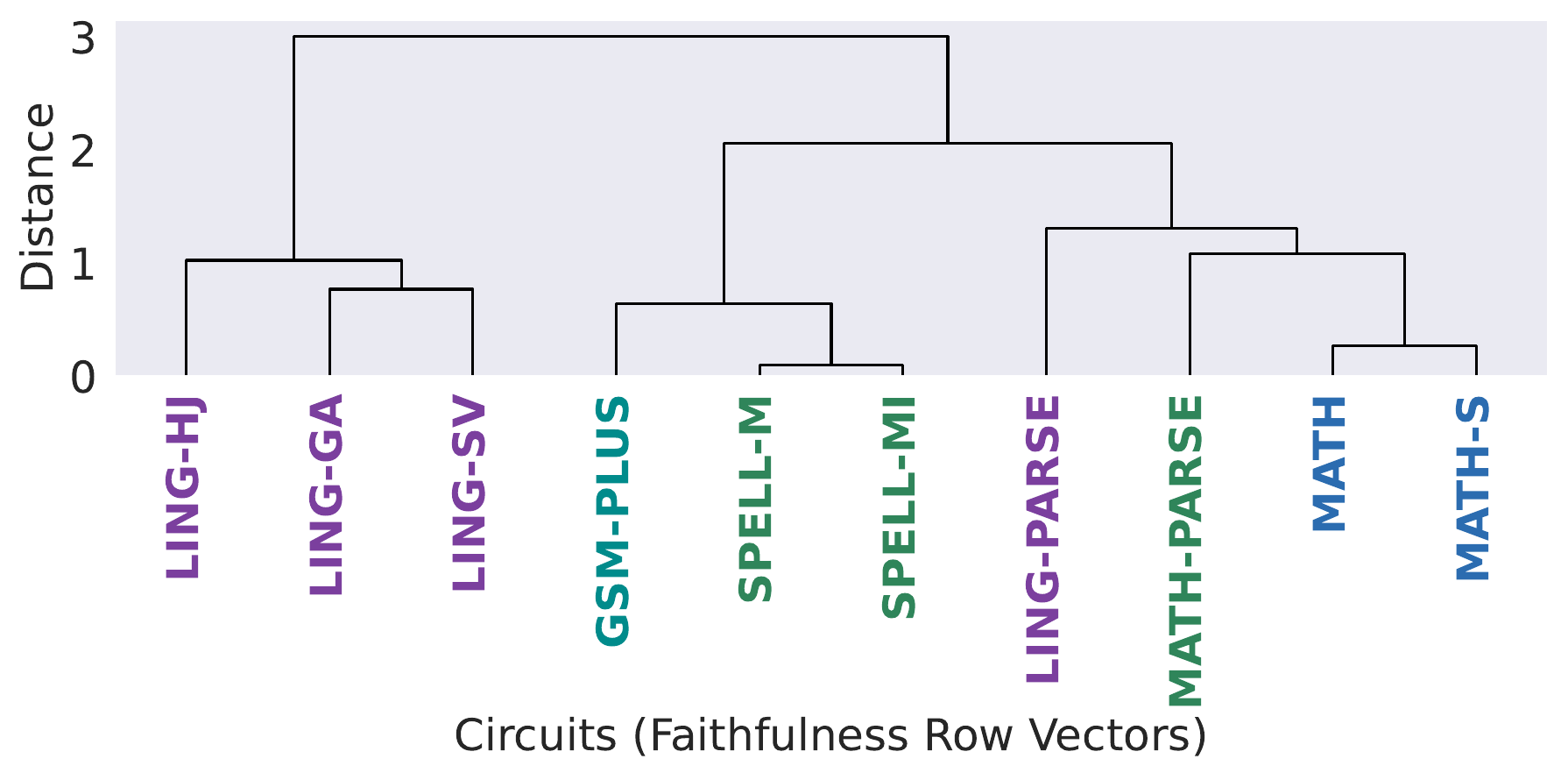}
  \caption{Hierarchical clustering of task circuits based on cross-task Faithfulness transfer profiles under the top-$2.5\%$ circuit definition. Top: Qwen3-8B. Bottom: Llama-3.1-8B.}
  \label{fig:qwen3-8-dendrogram}
\end{figure}

\subsection{Models}
\label{subsec:model}
In our experiments, we use Qwen3-8B, Qwen3-4B, Qwen3-1.7B~\cite{yang2025qwen3technicalreport}, Llama-3.1-8B~\cite{grattafiori2024llama3herdmodels}, and Gemma-2-2B (\texttt{google/gemma-2-2b})~\cite{gemmateam2024gemma2improvingopen}.
We include the Qwen3 series as a representative same-family set to examine how the results vary with model size, and additionally include Llama-3.1-8B and Gemma-2-2B to assess whether the observed patterns are consistent across model families.
Further details on the models are provided in Appendix~\ref{sec:model-detail}.
Table~\ref{tab:model-accuracy} shows model performance on each task, and models are generally capable of achieving high accuracy on the target tasks.
One concern is their relatively low performance on \textsc{GSM-Plus}, whereas this task is not crucial for our central research question and should just be regarded as a relevant reference.

\section{Experimental Results}
\label{sec:result}
This section shows results for Qwen3-8B and Llama-3.1-8B using the top 2.5\% of the circuit edges. 
Section~\ref{sec:analysis} shows the results for different circuit sizes, and Appendix~\ref{sec:appendix-other-models} shows the results for other models.

\paragraph{IoU.}
Figure~\ref{fig:npmi-faithfulness-all} (left) shows cross-task IoU scores.  
The scores are generally low, but those for the \textsc{Math} and \textsc{Math-s} pair and the \textsc{Spell-m} and \textsc{Spell-mi} pair are relatively high. This suggests that circuit overlap is affected by task characteristics rather than by the superficial input differences (e.g., Arabic numerals versus spelled-out expressions).
At the same time, the circuits for \textsc{Math} and \textsc{Spell} do not show remarkable similarity, indicating that the circuit reflects the task, rather than mere input similarity.

\paragraph{Faithfulness.}
Figure~\ref{fig:npmi-faithfulness-all} (right) shows cross-task Faithfulness scores. Each cell indicates how well the circuit extracted for the task in the row can solve the task in the column. 
The baseline-normalized values before diagonal display normalization are listed in Appendix~\ref{sec:appendix-other-models}.
Compared to IoU, task-dependent variations are more clearly observed in Faithfulness scores.
For example, arithmetic circuits of \textsc{Math} and \textsc{Math-s} show similar cross-task Faithfulness patterns, consistent with their high IoU. 
Similar tasks tend to have a relatively high faithfulness scores, for example, among expression-conversion tasks of \textsc{Spell-m} and \textsc{Spell-mi} and among the linguistic tasks of \textsc{Ling-SV}, \textsc{Ling-GA}, and \textsc{Ling-HJ}.
 \textsc{GSM-Plus} shows somewhat intermediate patterns between those of the calculation and expression-conversion circuits. 
 The notable divergence is observed for \textsc{Math-Parse} and \textsc{Ling-Parse}, where the low scores in their intersection indicate that the math/ling parse tasks cannot be solved by their counterparts.

\paragraph{Hierarchical clustering.}
To facilitate the interpretation of cross-task Faithfulness patterns, we also perform hierarchical clustering, using each row of the matrix as the vector for each circuit (Figure~\ref{fig:qwen3-8-dendrogram}). 
Each vector summarizes how well a source circuit solves all target tasks. 
We compare task clustering across the five evaluated models using the top-$2.5\%$ circuits.
For each model, we apply Ward linkage with Euclidean distance to the source-circuit row vectors of the diagonal-self-normalized cross-task faithfulness matrix, as in the main analysis.
For all model pairs, we compare partitions obtained by cutting the dendrograms into $K=2,\ldots,6$ clusters using the Adjusted Rand Index (ARI).
The ARI of 0 indicates no correlation, and 1 indicates perfect agreement.
Table~\ref{tab:clustering-ari} shows the statistics of ARI scores, yielding reasonably high agreement around 0.4--0.6.
This confirms that the cross-task circuit dependencies are generally stable across different models.

\paragraph{Language-arithmetic dissociation.}
The results reveal a general dissociation trend between language and arithmetic circuits. 
That is, circuits extracted from arithmetic tasks (\textsc{Math}, \textsc{Math-s}, \textsc{Math-Parse}, \textsc{GSM-Plus}) show limited overlap and low cross-task faithfulness with those from linguistic tasks (\textsc{Ling-SV}, \textsc{Ling-GA}, \textsc{Ling-HJ}, \textsc{Ling-Parse}). 
More nuanced patterns are observed in the \textsc{Spell} tasks and the \textsc{Math-Parse} task; we discuss their characteristics in Section~\ref{sec:discussion}.

\begin{table}[t]
  \centering
  \small
  \begin{tabular}{crrrr}
    \toprule
    $K$ & Mean ARI & Median ARI & Min & Max \\
    \midrule
    2 & 0.628 & 0.597 & 0.294 & 1.000 \\
    3 & 0.580 & 0.570 & 0.247 & 1.000 \\
    4 & 0.452 & 0.400 & 0.203 & 0.743 \\
    5 & 0.589 & 0.371 & 0.371 & 1.000 \\
    6 & 0.550 & 0.438 & 0.325 & 1.000 \\
    \bottomrule
  \end{tabular}
  \caption{Adjusted Rand Index (ARI) across all ten model pairs for partitions of the top-$2.5\%$ faithfulness dendrograms into $K$ clusters.}
  \label{tab:clustering-ari}
\end{table}

\section{Analysis}
\label{sec:analysis}

\begin{figure}[t]
  \centering

  \begin{subfigure}[t]{\linewidth}
    \centering
    \includegraphics[width=0.9\linewidth]{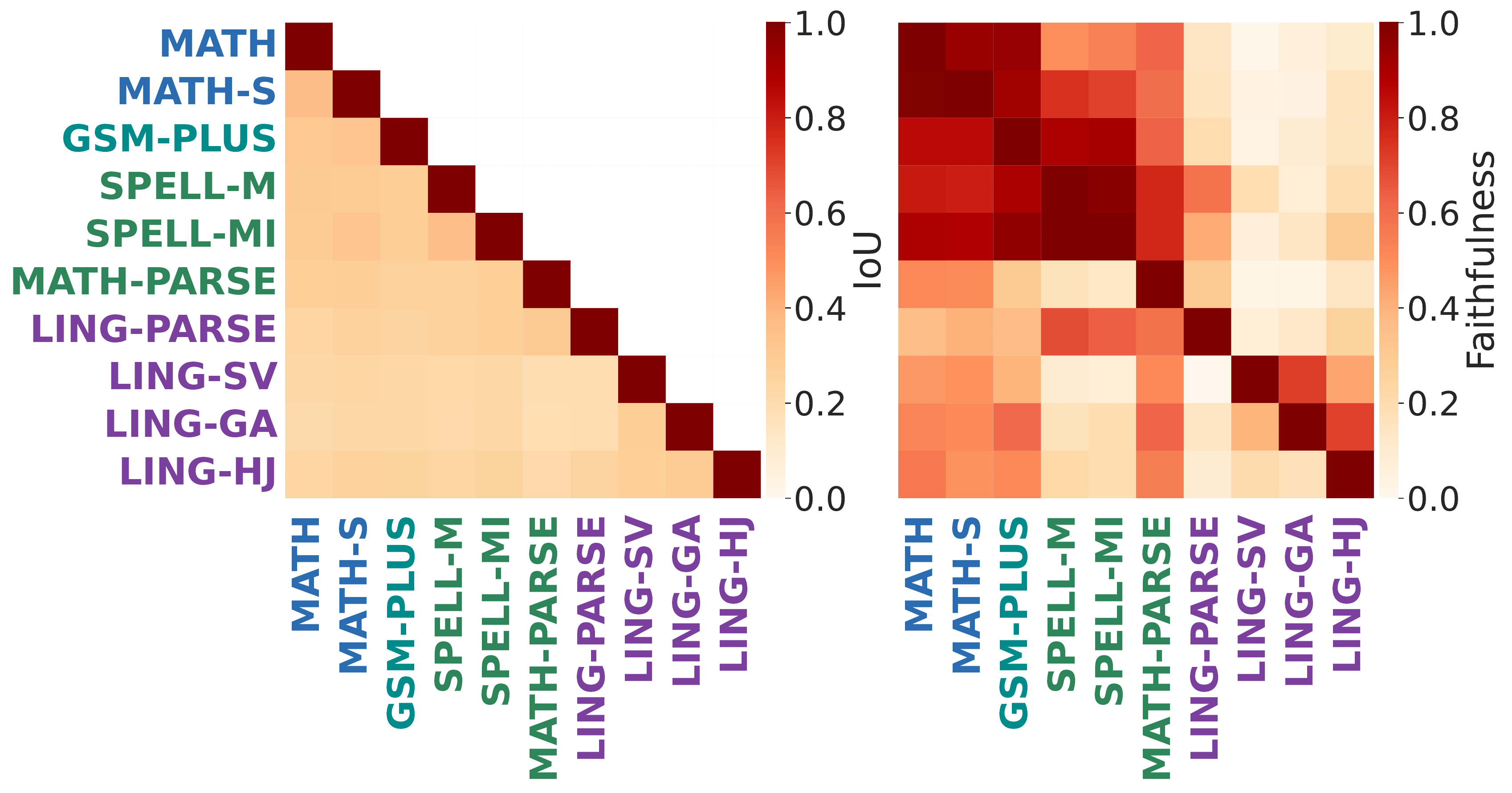}
    \caption{Cross-task IoU and Faithfulness scores, where rows denote source-circuit tasks and columns denote target tasks.}
  \end{subfigure}
    \begin{subfigure}[t]{\linewidth}
    \centering
  \includegraphics[width=\linewidth]{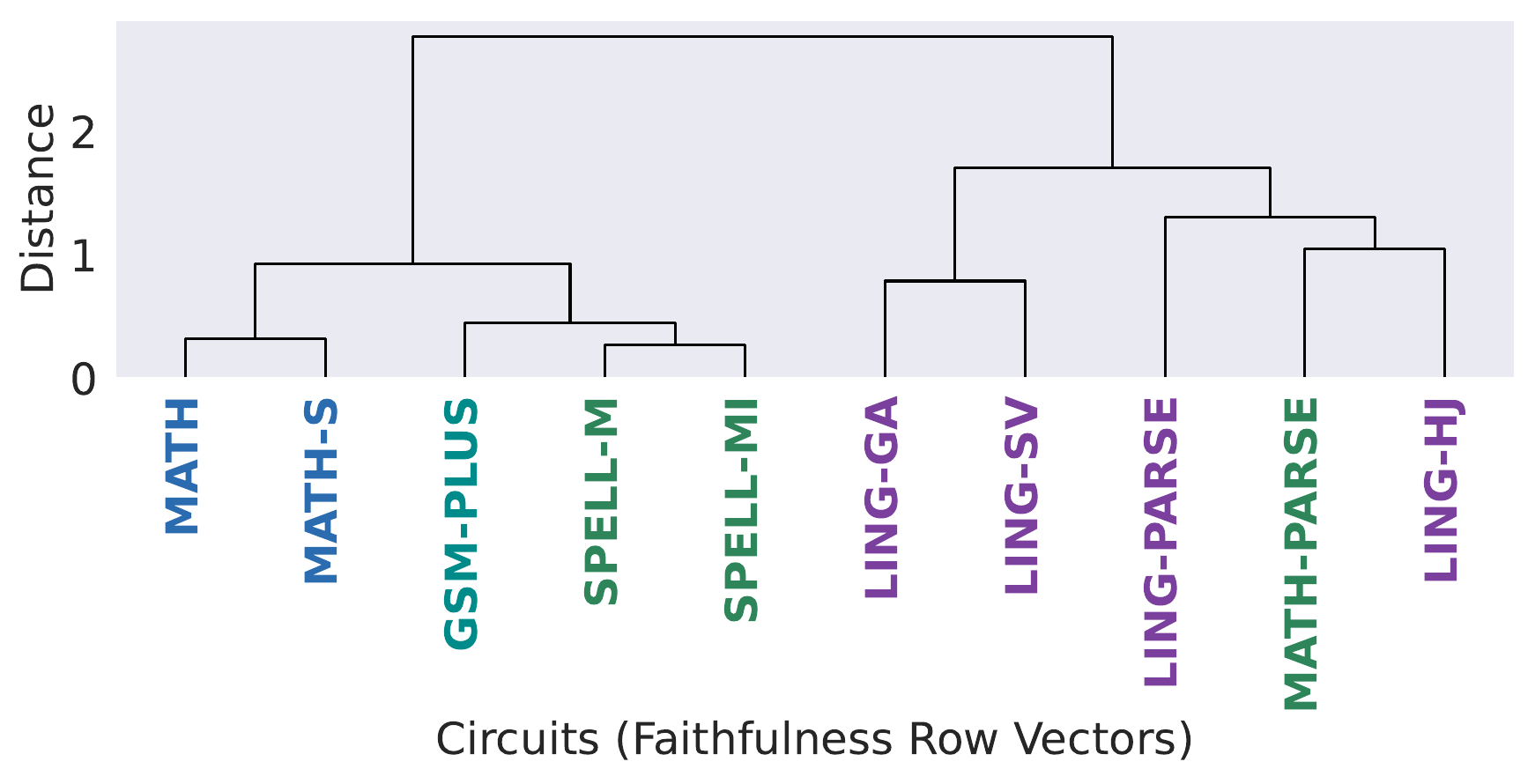}
      \caption{Hierarchical clustering of task circuits based on cross-task Faithfulness transfer profiles.}
  \end{subfigure}
  \caption{Results for Qwen3-8B under the top-$5.0\%$ circuit definition.}
  
  \vspace{1em}
  \label{fig:qwen3-8-5.0}
  
\end{figure}

\begin{figure}[t]
  \centering

  \begin{subfigure}[t]{\linewidth}
    \centering
    \includegraphics[width=\linewidth]{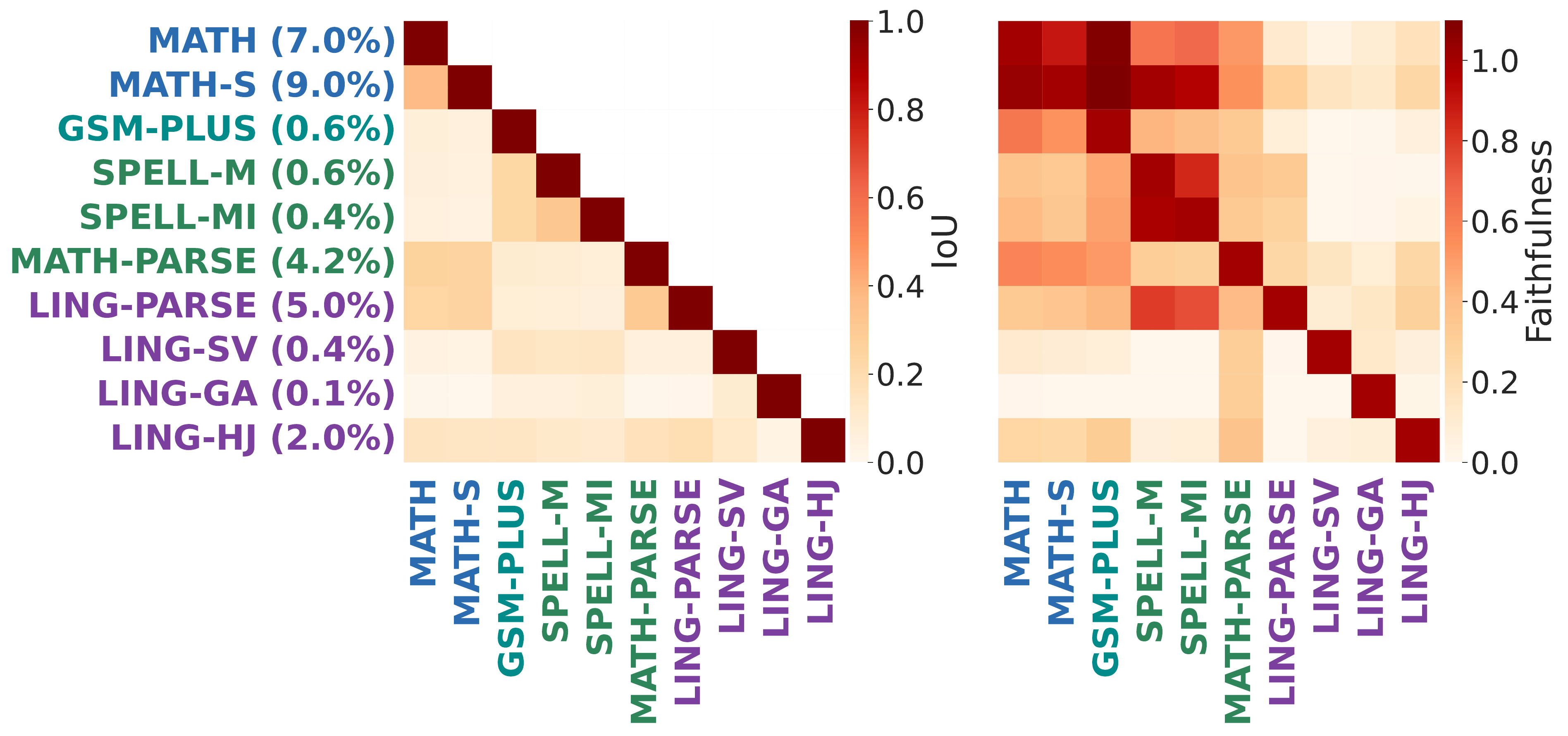}
    \caption{Cross-task IoU and Faithfulness scores, where rows denote source-circuit tasks and columns denote target tasks.}
  \end{subfigure}
  \begin{subfigure}[t]{\linewidth}
    \centering
  \includegraphics[width=\linewidth]{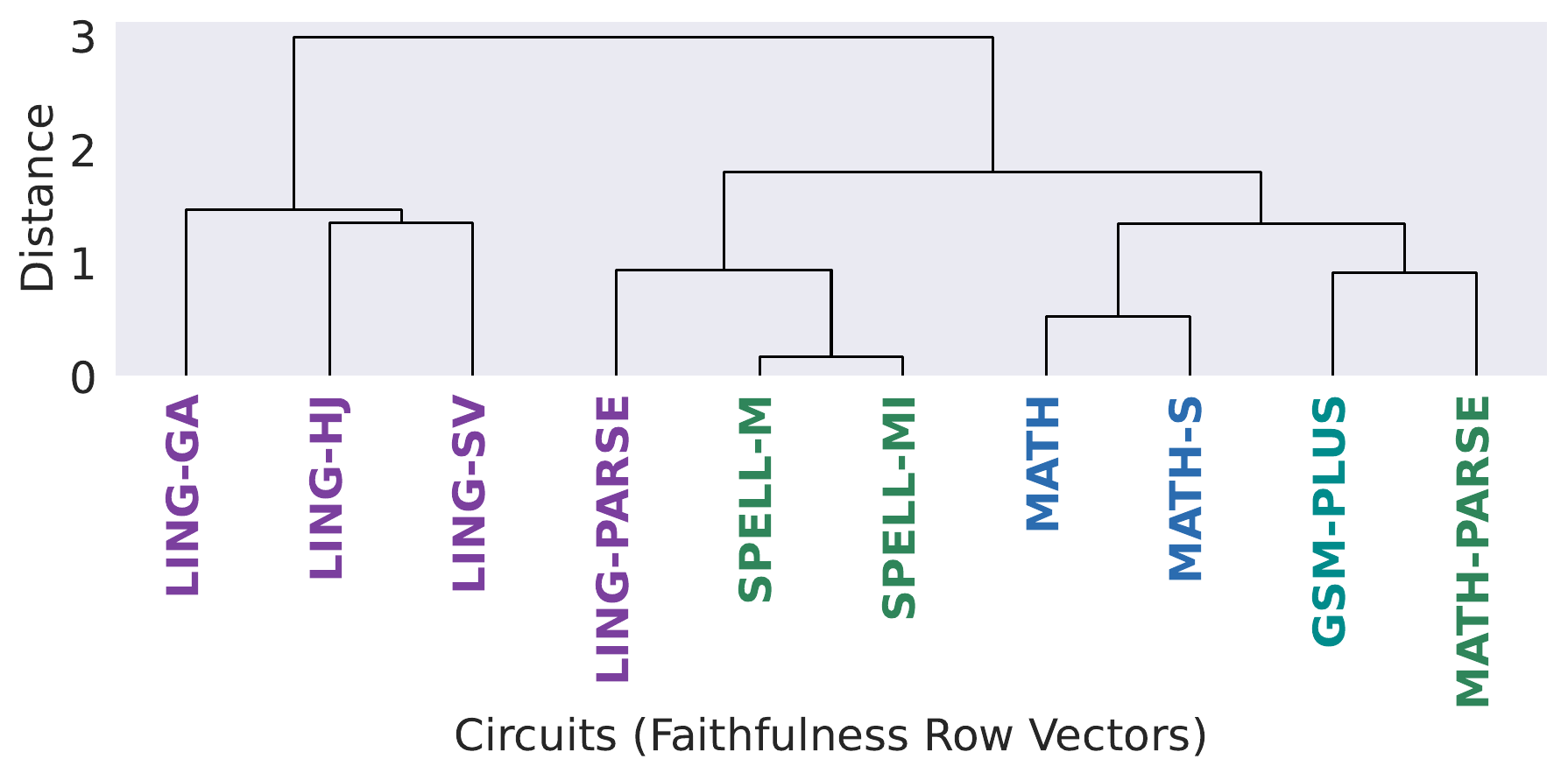}
  \caption{Hierarchical clustering of task circuits based on cross-task Faithfulness transfer profiles.}
    \end{subfigure}
  \caption{Results for Qwen3-8B under the adaptive circuit size with the self-faithfulness threshold of $0.85$.}
  
  \vspace{1em}
  \label{fig:qwen3-8-self}
  
\end{figure}

\subsection{Different Circuit Size}
To investigate whether the obtained results depend on the circuit size threshold, we also conducted experiments using the top-$5.0\%$ circuit, instead of 2.5\%.
Figure~\ref{fig:qwen3-8-5.0} shows the results for Qwen3-8B as a representative example. The broad dissociation pattern remains similar to that under the top-$2.5\%$ definition.

\subsection{Adaptive Circuit Size}
\label{subsec:adaptive_circuit}
As the fixed-size circuit definitions may not reflect the actual minimal circuitry required for each task, we also consider an adaptive circuit size, i.e., the smallest circuit whose self-faithfulness exceeds a certain threshold.
Here, we examine $0.85$ as the threshold for self-faithfulness to define the adaptive circuit.
Figure~\ref{fig:qwen3-8-self} shows the results for Qwen3-8B with adaptive circuit size. 
The broad separation between arithmetic and linguistic tasks remains clear under this definition.
This setting provides rather clearer inter-task IoU, while cross-task Faithfulness shows reduced values, especially for the circuits of \textsc{Spell-m} and \textsc{Spell-mi}. 
This likely reflects that 2.5\% fixed-size circuits for \textsc{Spell-m} and \textsc{Spell-mi} may include extra edges beyond spelling.
Indeed, in Figure~\ref{fig:llama-3.1-8b-npmi-faithfulness}, the \textsc{Spell-m} circuits could solve \textsc{Math} tasks reasonably, and this is mitigated in the adaptive circuit definition, which likely removes these extra edges, leading to a more accurate representation of the task-specific circuitry.
This setting alters some hierarchical clustering results (e.g., similarity between \textsc{Spell-m} and \textsc{GSM-Plus}). 
However, the general dissociation between arithmetic and language circuits remains intact, or rather becomes much clearer.

\section{Discussion}
\label{sec:discussion}

Our results suggest that arithmetic circuits are generally dissociated from those supporting the purely linguistic tasks.
This is in line with the general view of neural dissociation between language and arithmetic processing in human studies~\cite{varley2005agrammatic,friedrich2009mathematical,maruyama2012cortical,amalric2018cortical,Fedorenko2024-dd}.
In other words, the potential interactions between language and arithmetic, suggested by~\citet{Nakai2018}, cannot clearly be observed for LLMs.
While \textsc{Math-Parse} and \textsc{Ling-Parse} circuits are sometimes close in hierarchical clustering results, their cross-task faithfulness is quite limited (Figure~\ref{fig:npmi-faithfulness-all}), in other words, the arithmetic/language parsing circuit cannot solve the counterpart parsing task.
Thus, it is difficult to say that there is a shared circuit between arithmetic and language processing, even in settings that maximally emphasize their common characteristics.

Notably, even though the hierarchical clustering results are generally stable across different models and settings, \textsc{Math-Parse} circuit is typically ``floating'' in the hierarchical clustering results, i.e., it does not consistently cluster with either the broader arithmetic or linguistic task groups.
In addition, \textsc{Math} circuits cannot perfectly solve the \textsc{Math-Parse} task, and vice versa, suggesting that the \textsc{Math-Parse} circuit found in this study is not involved in the general arithmetic processing.
One suspicion is that LMs do not clearly recognize that parsing arithmetic expressions is the essential subtask of arithmetic, i.e., there is no rigid circuit for this parsing subtask, and perhaps solve arithmetic problems with a different strategy internally from humans, e.g., compute multiple terms at once, potentially with memorization or with a bag of heuristics~\cite{nikankin2025arithmetic}, rather than following a step-by-step parsing-based approach.
This offers a hypothetical reason why language and arithmetic circuits are dissociated: LMs solve arithmetic differently from humans, and perhaps there may no longer be a perception that language and arithmetic share fundamental structural characteristics.

Another notable point is that the structural overlap (IoU) of circuits is sometimes altered by the functional overlap analysis (Faithfulness).
A plausible interpretation is that a relatively small set of edges plays a substantial role in supporting several tasks, and thus merely counting the number of overlapping edges may underweight their importance.
Such a pattern is consistent with recent discussions of circuit polysemy and with the possibility that some edges behave like shared high-impact routing components.
Under this view, low structural overlap does not necessarily imply strong mechanistic independence.
\section{Conclusion}
\label{sec:conclusion}
We analyzed the characteristics of task circuits using controlled arithmetic-related and linguistic tasks, across five LLMs. 
The extracted arithmetic and linguistic circuits are largely dissociated, echoing the neuro-linguistic findings in humans.
Nevertheless, human experiments also suggested some interaction between these two domains, particularly stemming from the shared structural and recursive nature of language and arithmetic expressions, and LLMs do not exhibit such interactions.
The circuits for \textsc{Ling-Parse} and \textsc{Math-Parse}, where the task is structural analysis of input sentences and arithmetic expressions, show limited mutual transfer. 
We tentatively conclude that there is no clear evidence that LLMs equip a shared circuit for primitive language and arithmetic processing. 

\section*{Limitations}
Our results might be biased by the specific configurations we adopted, and further exploration is needed to enhance the generality of the findings.
First, we only used the EAP-IG method for circuit extraction. 
Exploring and comparing other circuit extraction methods should enhance the generality of our observations, while we examined different hyperparameters, such as top-$2.5\%$, top-$5.0\%$, and adaptive circuits.
Relatedly, the main experiments target the first answer token alone, while some tasks, such as spelling-out ones, are defined with sequence output. 
Appendix~\ref{sec:sequence-evaluation} preliminarily examines this bias, and integrating sequence-level generation for circuit extraction and evaluations will further clarify the findings. 
Second, the task and evaluation settings might be biased; for example, all prompts are in the English language, and the linguistic tasks are somewhat heterogeneous. 
Generally low accuracy on \textsc{GSM-Plus} may also weaken interpretations, regarding the distinction between mathematical reasoning and pure arithmetic processing. 
Third, we evaluate a finite set of open-weight models and training regimes. 
Lastly, the connection between our LLM results and neurolinguistic findings on the human side should be further elaborated, which opens the question of humanlikeness of LLMs from the neural network modularity perspectives.

\section*{Ethical considerations}
This work analyzes existing large language models and does not involve human-subject experiments. Part of our evaluation data was newly created using hand-written templates, and is fully synthetic, i.e., it does not contain text sourced from individuals or potentially sensitive personal information.
We used AI assistants for coding assistance in building the experimental pipeline and for adjusting wording. The authors are responsible for verifying the implementation, citations, and scientific claims.

\bibliography{custom}

\clearpage
\appendix
\section{Experimental Settings}
\label{sec:appendix-exp-settings}

\paragraph{Circuit extraction.}
We extract task circuits using Edge Attribution Patching with Integrated Gradients (EAP-IG).
For each \emph{(model, task)} pair, we run attribution to obtain an importance score for every edge in the analysis graph,
and evaluate three circuit definitions: the subsets of edges whose absolute attribution scores fall within the top 2.5\% and top 5.0\% of all edges, and an adaptive circuit defined as the smallest edge set whose
self-faithfulness exceeds 0.85. Unless otherwise noted, we use IG steps = 5, run on a single GPU in float16.
On a single NVIDIA RTX PRO 6000 Blackwell Max-Q Workstation Edition (96GB VRAM), this step takes approximately
5 GPU-hours per (model, task) pair.

\paragraph{Cross-task evaluation.}
To measure functional overlap between tasks, we evaluate how well the circuit extracted from task $B$ supports task $A$.
For each ordered pair $(A,B)$, we apply the circuit of $B$ as a structural mask and run edge patching on task $A$ using
the clean/corrupted pair $(x^{A},\tilde{x}^{A})$:
we keep \emph{clean} edge activations on the edges in the circuit of $B$ and patch \emph{all remaining edges} with the
corresponding activations from the corrupted run on $\tilde{x}^{A}$ (cf.\ Eq.~\ref{eq:mixed_edge_acts}).
We then evaluate the task-$A$ metric under this mixed forward pass and record the resulting score together with the
clean and corrupted baselines. 
On the same GPU, computing a full cross-task matrix for one model takes approximately 20 GPU-hours.

\paragraph{Heatmaps and hierarchical clustering.}
For each circuit definition, we visualize (i) structural overlap (IoU) and (ii) diagonal-normalized
cross-task faithfulness as heatmaps. In the Faithfulness heatmaps, rows correspond to source-circuit tasks and columns correspond to target tasks.
For task grouping analysis, we perform hierarchical clustering on the row vectors of the IoU or displayed normalized faithfulness
matrices using agglomerative clustering (Ward linkage with Euclidean distance), and export dendrograms. For Faithfulness, this groups tasks by the transfer profiles of their extracted circuits.

\section{Tasks}
\label{sec:tasks-detail}
Task definitions are provided in Section~\ref{subsec:task} (Table~\ref{tab:datalist}).
Here, we provide prompt templates and clean/corrupted examples (Table~\ref{tab:appendix_clean_corrupted_loss_template}), followed by details of dataset construction and pairing.

\newcolumntype{L}[1]{>{\raggedright\arraybackslash}p{#1}}
\newcolumntype{T}[1]{>{\raggedright\arraybackslash\ttfamily}p{#1}}

\begin{table*}[t]
  \centering
  \footnotesize
  \tabcolsep=0.10cm
  \renewcommand{\arraystretch}{1.15}

  \begin{tabular}{@{}L{3.0cm} T{6.3cm} L{4.6cm} c@{}}
    \toprule
    \textbf{Task}
      & \multicolumn{1}{c}{\textbf{Input}}
      & \multicolumn{1}{c}{\textbf{Expected Output}}
      & \textbf{Loss} \\
    \cmidrule(lr){2-2}\cmidrule(lr){3-3}
      & \textbf{Clean}     & \textbf{Clean}     & \\
      & \textbf{Corrupted} & \textbf{Corrupted} & \\
    \midrule

    Arithmetic calculation\newline(\textsc{Math})
      & The answer to 2 * 12 + 3 = is\newline
        The answer to 13 - 4 + 9 = is
      & 27\newline
        18
      & logit\_diff \\
    Spelled-out arithmetic calculation\newline(\textsc{Math-s})
      & The answer to two times twelve plus three equals is\newline
        The answer to thirteen minus four plus nine equals is
      & twenty-seven\newline\newline
        eighteen
      & logit\_diff \\
    Math Word Problem\newline(\textsc{GSM-Plus})
      & Q: Janet's ducks lay 16 eggs per day...\newline
        Q: Janet's ducks lay 10 eggs per day...
      & 18\newline
        27
      & logit\_diff \\

    Arithmetic expression spelling-out\newline(\textsc{Spell-m})
      & The spelled-out form of 2 * 12 + 3 is\newline
        The spelled-out form of 13 - 4 + 9 is
      & two times twelve plus three\newline
        thirteen minus four plus nine
      & logit\_diff \\

    Inverse of \textsc{Spell-m}\newline(\textsc{Spell-mi})
      & The numerical formula for two times twelve plus three is\newline
        The numerical formula for thirteen minus four plus nine is
      & \texttt{2 * 12 + 3}\newline\newline
        \texttt{13 - 4 + 9}
      & logit\_diff \\

    Arithmetic expression parsing\newline(\textsc{Math-Parse})
      & The first term to calculate in 6 + 2 * 5 is\newline
        The first term to calculate in 6 * 2 + 5 is
      & \texttt{2 * 5}\newline
        \texttt{6 * 2}
      & logit\_diff \\

    Sentence parsing\newline(\textsc{Ling-Parse})
      & The head word of the lowest constituent of the sentence 'many students study' is\newline
        The head word of the lowest constituent of the sentence 'students often study' is
      & students\newline\newline
        study
      & logit\_diff \\

    Subject--Verb Agreement\newline(\textsc{Ling-SV})
      & The keys on the cabinet\newline
        The key on the cabinet
      & are\newline
        is
      & logit\_diff \\

    Gendered Pronoun Agreement\newline(\textsc{Ling-GA})
      & The mother cried because\newline
        The father cried because
      & she\newline
        he
      & logit\_diff \\

    Hypernym Judgment\newline(\textsc{Ling-HJ})
      & roses, and other\newline
        folk, and other
      & flower\newline
        music
      & prob\_diff \\
    \bottomrule
  \end{tabular}

  \caption{Example instances for each task in our controlled suite. For each task, we show a clean/corrupted prefix pair and the corresponding expected output, along with the loss used for evaluation (logit\_diff or prob\_diff).}
  \label{tab:appendix_clean_corrupted_loss_template}
\end{table*}

We apply tokenizer-dependent filters separately for each model family.
Within each pair, we match tokenized prompt lengths to preserve positional alignment during edge patching.
For first-token evaluation, we require the clean and corrupted target answer tokens to differ.

\paragraph{Arithmetic and expression tasks.}
\textsc{Math}, \textsc{Math-s}, \textsc{Spell-m}, and \textsc{Spell-mi} use a shared set of three-term expressions.
The first operand ranges from 0 to 20, the other two from 0 to 50, and the operators are $+$, $-$, and $\times$.
We exclude negative answers and answers with absolute value at least 200.
Each clean expression is paired with a distinct corrupted expression that has the same tokenized prompt length across all four tasks.
For \textsc{Math-Parse}, we generate expressions from three numbers between 0 and 50, one multiplication, and one addition or subtraction.
Each pair uses the same numbers but places the multiplication term on opposite sides of the expression; the target is the first multiplication subexpression.

\paragraph{Linguistic tasks.}
For \textsc{Ling-Parse}, we generate short adjective--noun--verb, noun--adverb--verb, and noun--verb--adverb sentences from vocabulary lists.
Paired sentences share a noun and verb but differ in whether the lowest constituent has a noun or verb head; we also require single-token answers.
\textsc{Ling-SV} uses singular and plural examples from 15 subject--verb agreement templates, paired in both directions.
For \textsc{Ling-GA}, we swap masculine and feminine forms in profession-based gender pairs and use \textit{he} and \textit{she} as contrasting answers.
For \textsc{Ling-HJ}, we use filtered Battig category norms and valid hypernym forms.
We pair hyponyms from different semantic categories while matching hyponym and hypernym token lengths, allowing multiple valid hypernym forms.
For each hyponym, we specify a set of acceptable hypernym answers.
Given a clean/corrupted pair, let $H_c$ and $H_r$ be the respective answer sets.
For either prompt $x$ in the pair, the probability-difference metric (\texttt{prob\_diff}) is
\begin{equation}
m(x)=\sum_{w\in H_c}p(w\mid x)-\sum_{w\in H_r}p(w\mid x),
\end{equation}
where $p(w\mid x)$ denotes the model probability of the first token of candidate answer $w$ given $x$.
Thus, the metric compares the total probability assigned to the clean hyponym's valid hypernyms with that assigned to the corrupted hyponym's valid hypernyms.

\paragraph{Math word problems.}
\textsc{GSM-Plus} uses numerical-substitution examples from the test split of GSM-Plus.
We pair each seed question with its numerical-substitution variant.

\section{Model Detail}
\label{sec:model-detail}
The license of the used models is listed in \autoref{tab:licence}; all of them are used under their intended use.
All models were accessed using the Hugging Face toolkit~\cite{wolf-etal-2020-transformers}.

\begin{table*}[h]
    \small
    \centering
    \tabcolsep=0.1cm
    \begin{tabular}{llp{4cm}}
    \toprule
         Model&HuggingFace&License  \\
         \cmidrule(r){1-1}\cmidrule(r){2-2}\cmidrule(r){3-3}
         Llama-3.1-8B&\url{https://huggingface.co/meta-llama/Llama-3.1-8B}&Llama 3.1 Community License\\
         Qwen3-8B&\url{https://huggingface.co/Qwen/Qwen3-8B}&Apache license 2.0\\
         Qwen3-1.7B&\url{https://huggingface.co/Qwen/Qwen3-1.7B}&Apache license 2.0\\
         Qwen3-4B&\url{https://huggingface.co/Qwen/Qwen3-4B}&Apache license 2.0\\
         Gemma-2-2B&\url{https://huggingface.co/google/gemma-2-2b}&Gemma Terms of Use License\\
    \bottomrule
    \end{tabular}
    \caption{License of the models}
    \label{tab:licence}
\end{table*}

\section{Additional Results}
\label{sec:appendix-other-models}
\begin{figure}[t]
  \centering
  \begin{subfigure}[t]{\linewidth}
    \centering
    \includegraphics[width=\linewidth]{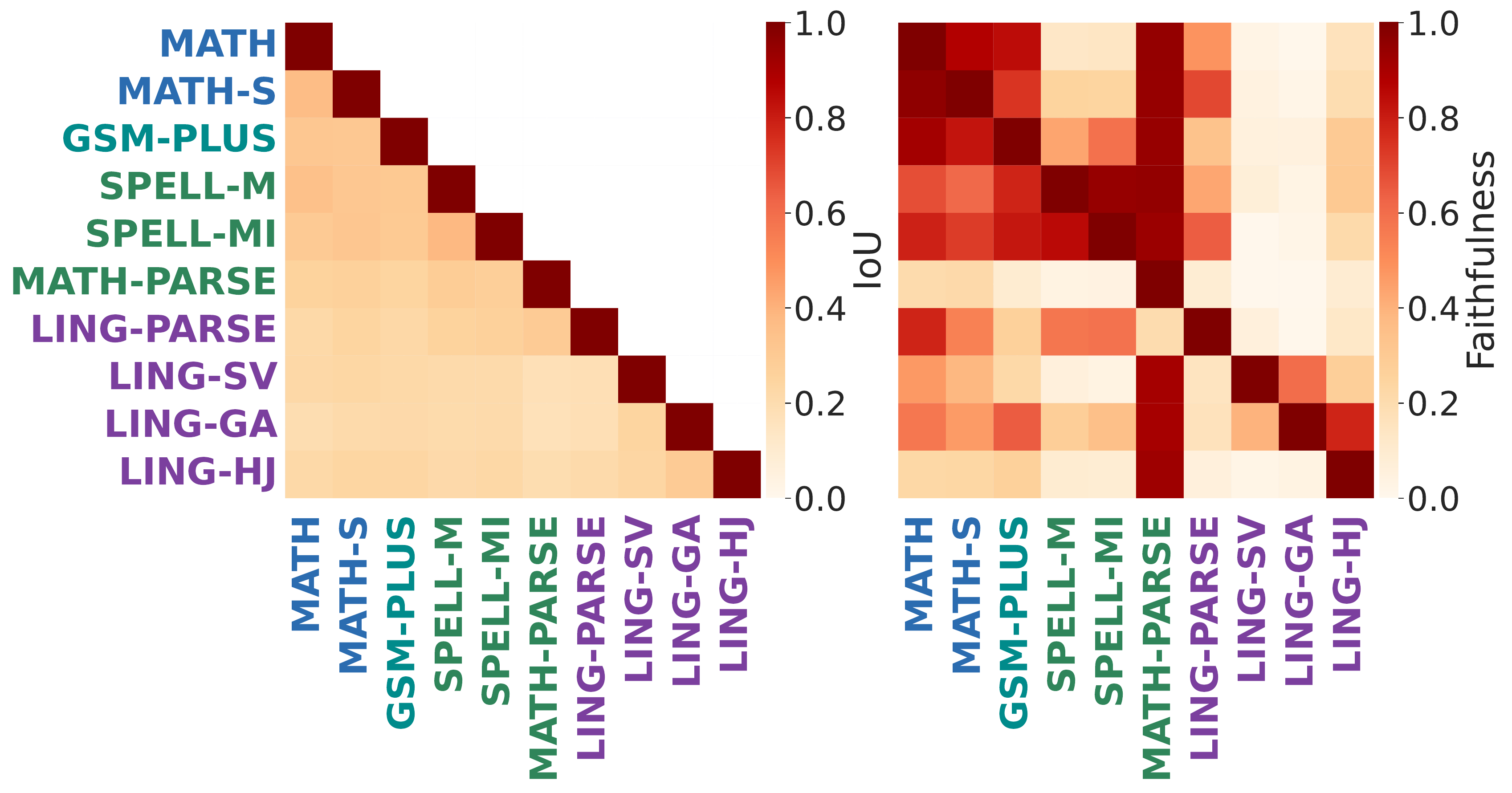}
    \caption{Qwen3-1.7B}
    \label{fig:qwen3-1.7-iou-faithfulness}
  \end{subfigure}

  \vspace{1em}

    \begin{subfigure}[t]{\linewidth}
    \centering
    \includegraphics[width=\linewidth]{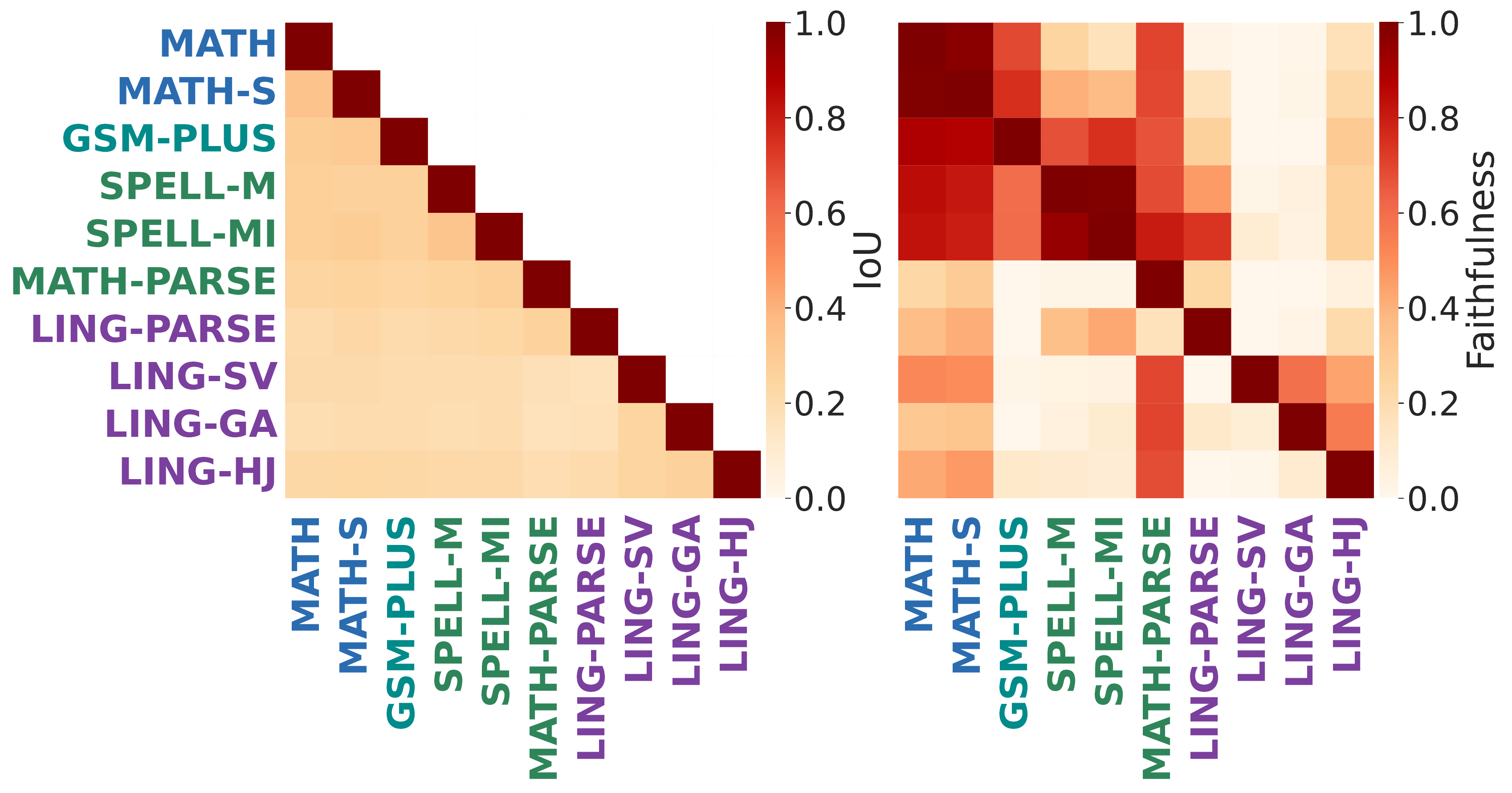}
    \caption{Qwen3-4B}
    \label{fig:qwen3-4-iou-faithfulness}
  \end{subfigure}

  \vspace{1em}
  
  \begin{subfigure}[t]{\linewidth}
    \centering
    \includegraphics[width=\linewidth]{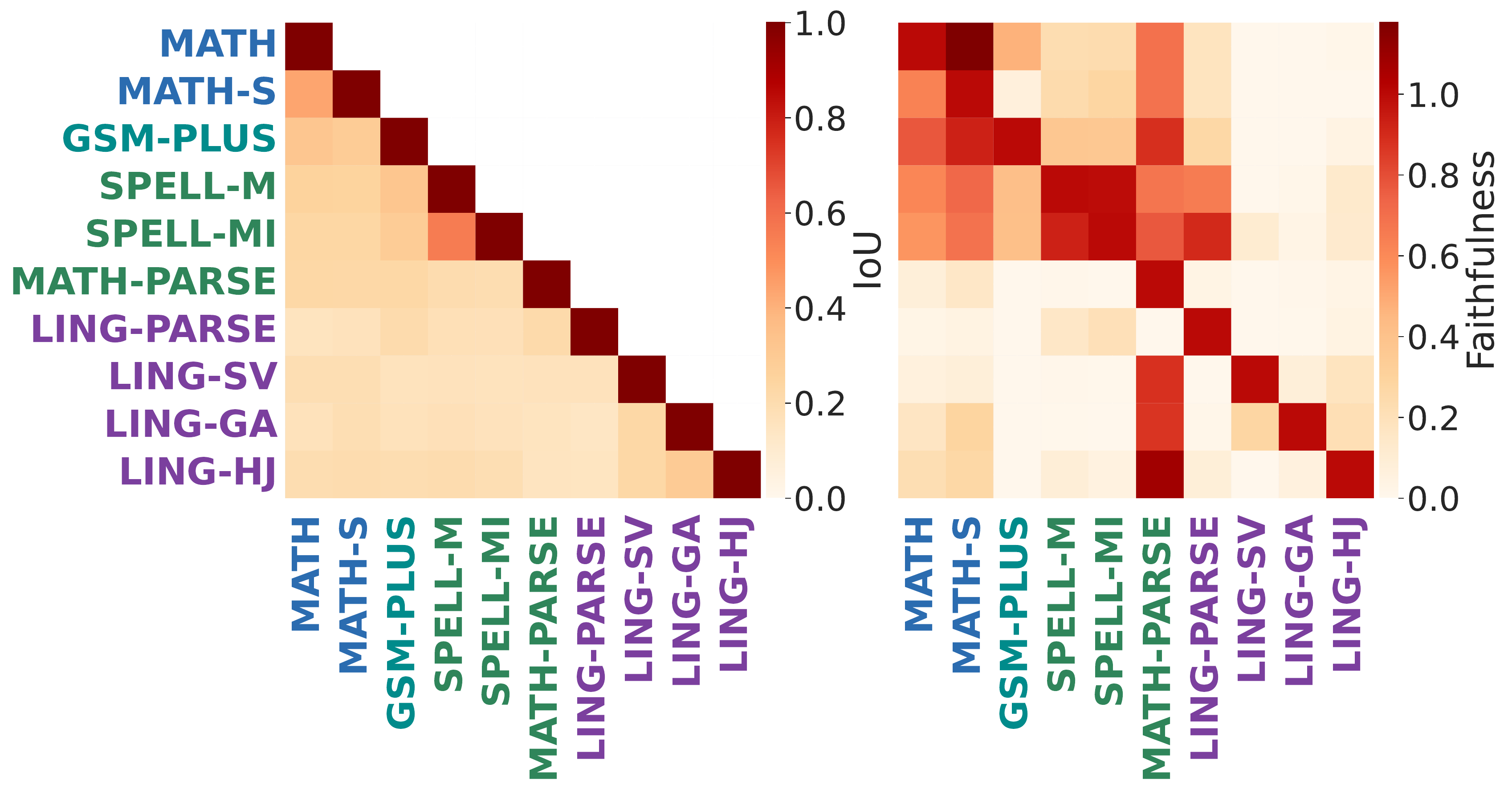}
    \caption{Gemma-2-2B}
    \label{fig:gemma-2-2b-iou-faithfulness}
  \end{subfigure}
  \caption{Left: IoU (structural overlap). Right: cross-task Faithfulness (functional overlap). For Faithfulness, rows denote the tasks from which circuits are extracted, and columns denote the target tasks solved with those circuits. Scores are normalized by the corresponding diagonal self-faithfulness value.}
  \label{fig:iou-faithfulness-other}
\end{figure}

\begin{figure}[t]
  \centering
  \begin{subfigure}[t]{\linewidth}
    \centering
    \includegraphics[width=\linewidth]{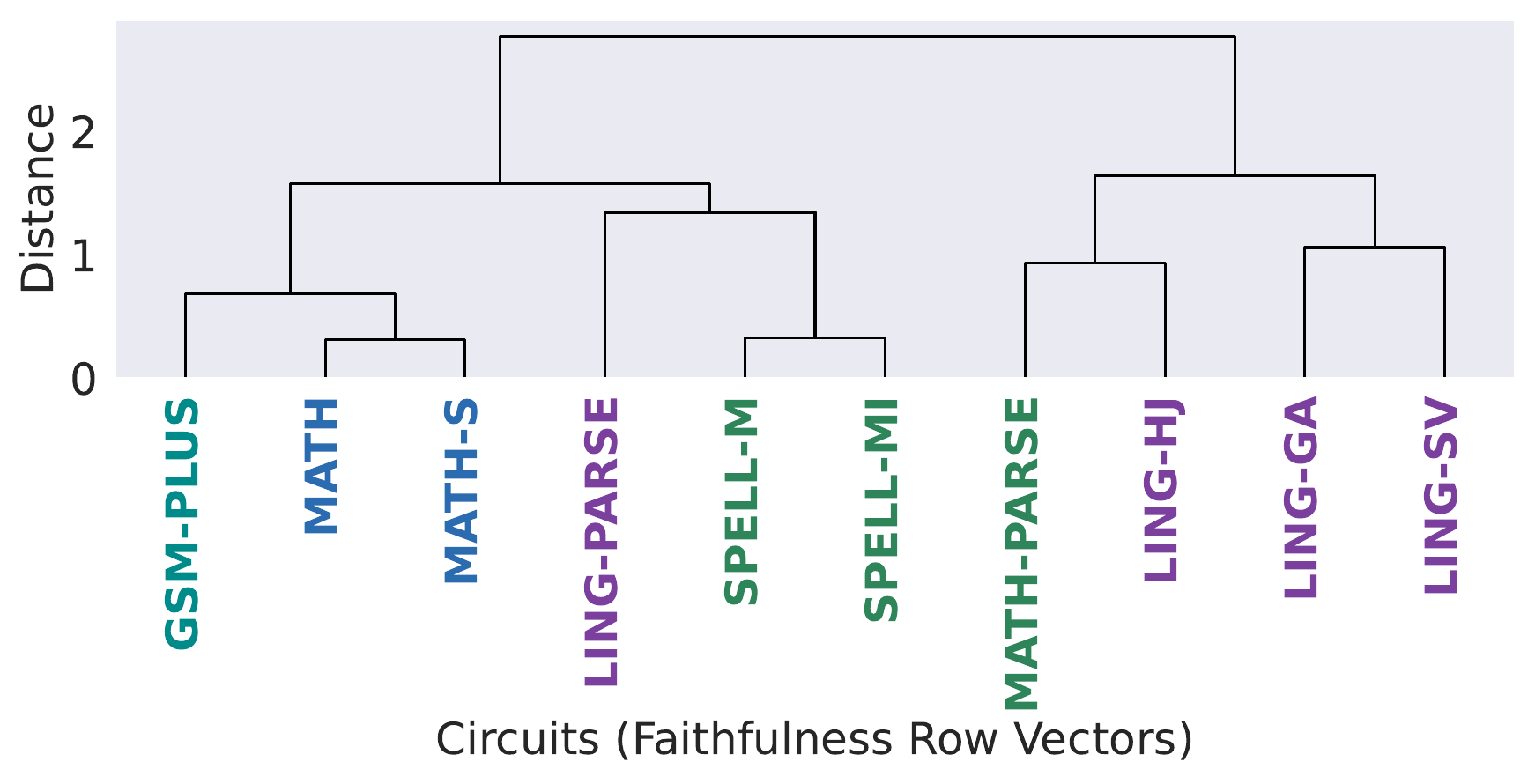}
    \caption{Qwen3-1.7B}
    \label{fig:qwen3-1.7-dendrogram}
  \end{subfigure}
  \vspace{1em}
    \begin{subfigure}[t]{\linewidth}
    \centering
    \includegraphics[width=\linewidth]{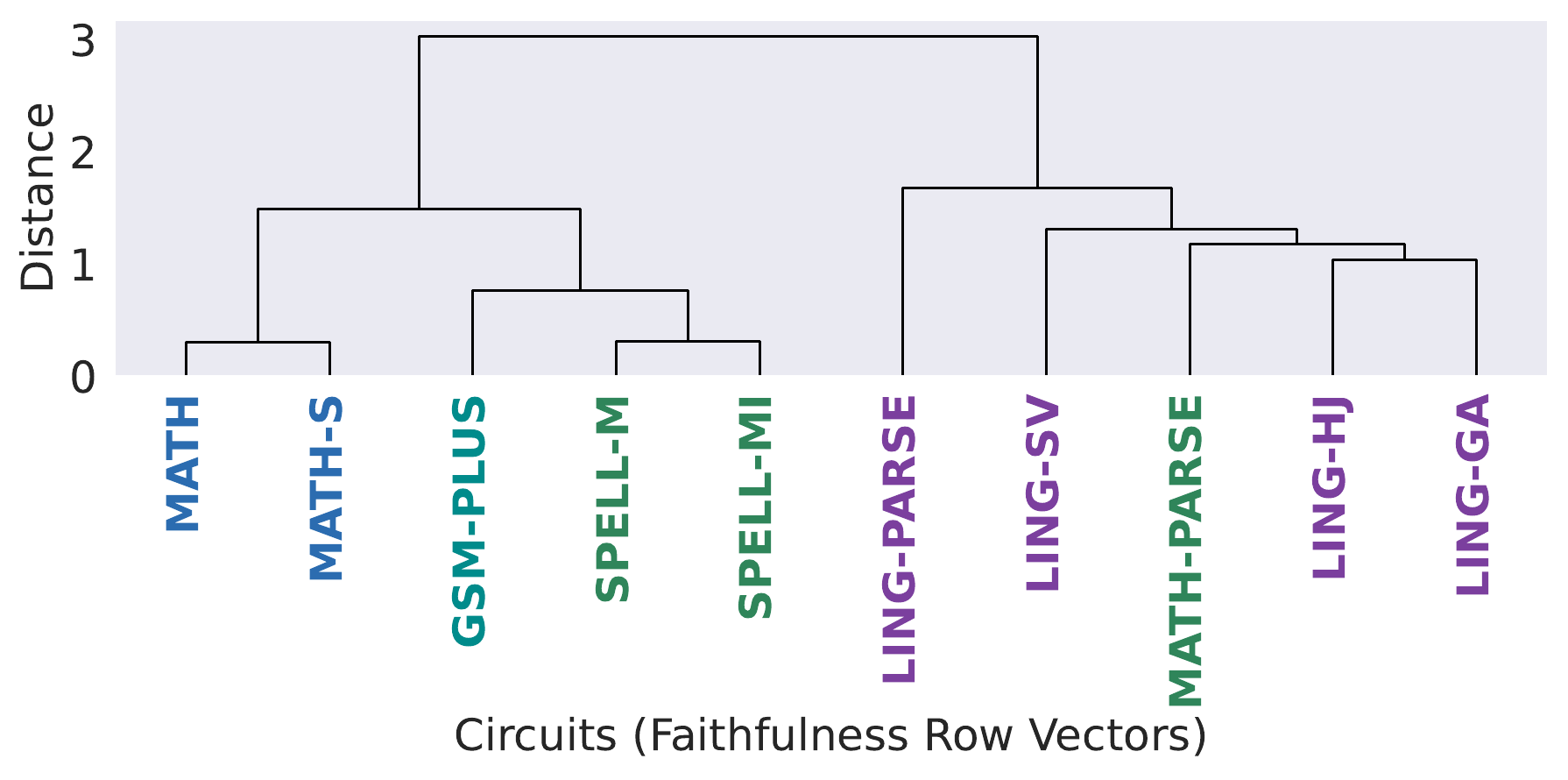}
    \caption{Qwen3-4B}
    \label{fig:qwen3-4-dendrogram}
  \end{subfigure}
  
  \vspace{1em}

    \begin{subfigure}[t]{\linewidth}
    \centering
    \includegraphics[width=\linewidth]{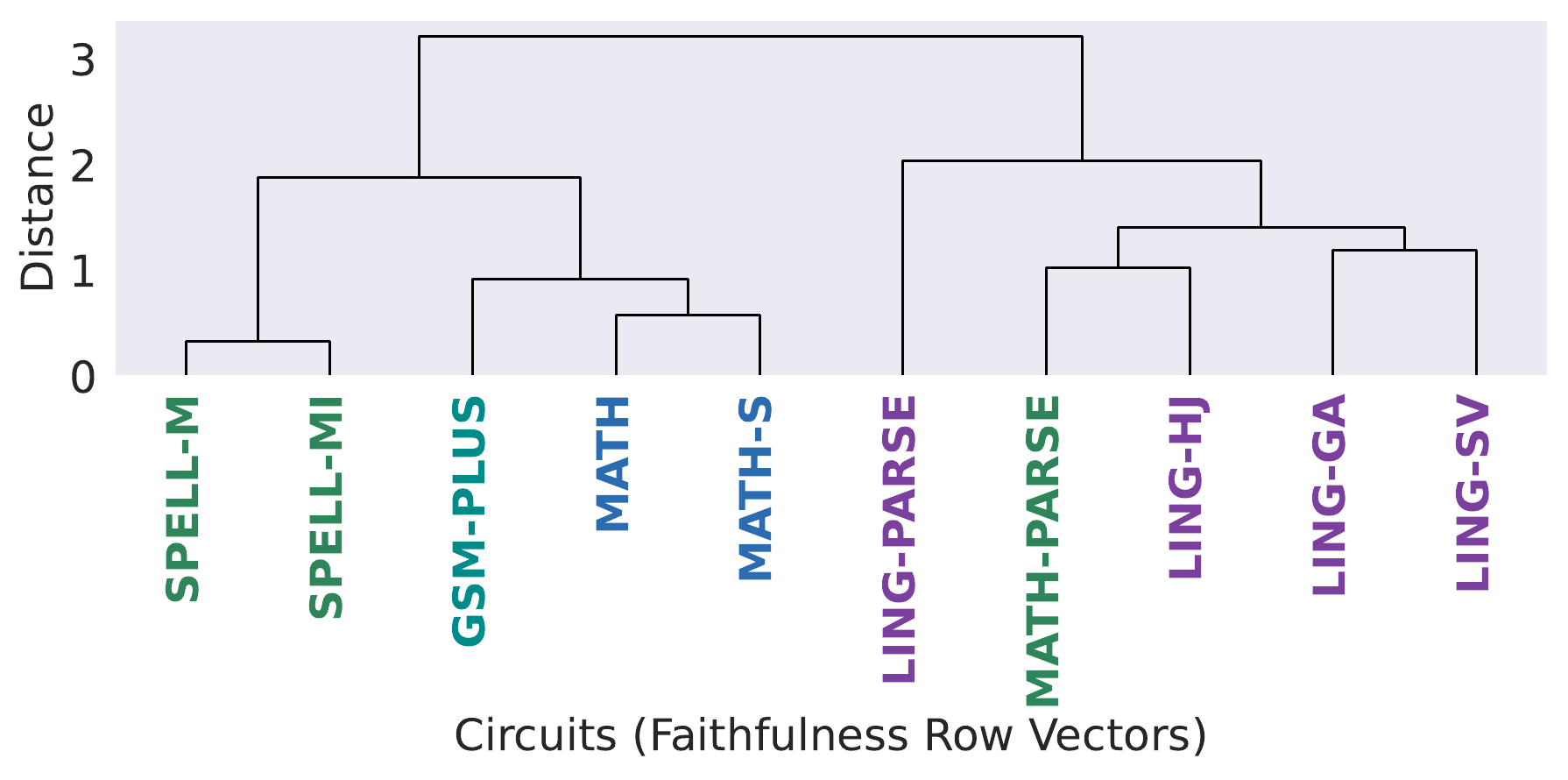}
    \caption{Gemma-2-2B}
    \label{fig:gemma-2-2b-dendrogram}
  \end{subfigure}
  \caption{Hierarchical clustering of task circuits based on cross-task Faithfulness transfer profiles. Top: Qwen3-1.7B. Middle: Qwen3-4B. Bottom: Gemma-2-2B.}
  \label{fig:dendrogram-other}
\end{figure}

We present results for additional models (Gemma-2-2B~\cite{gemmateam2024gemma2improvingopen}, Qwen3-1.7B, and Qwen3-4B) in Figure~\ref{fig:iou-faithfulness-other} and Figure~\ref{fig:dendrogram-other}.
The smaller Qwen3 models show several patterns consistent with the main results, including strong \textsc{Math}/\textsc{Math-s} transfer and limited bidirectional transfer between the parsing tasks.

Section~\ref{sec:result} summarizes cross-task patterns using heatmaps. To enable detailed inspection and precise comparison, we list the raw cross-task Faithfulness matrices for each model under the top-$2.5\%$ circuit definition in Tables~\ref{table:llama-3.1-8b-faithfulness}--~\ref{table:gemma-2-2b-faithfulness}. These tables report baseline-normalized faithfulness, not absolute logit/probability scores, before the column-wise normalization by diagonal self-faithfulness used in the displayed Faithfulness heatmaps. Rows correspond to source-circuit tasks and columns correspond to target tasks.

\begin{table*}[t]
  \centering
  \scriptsize
  \tabcolsep=0.1cm
  \begin{tabular}{lrrrrrrrrrr}
    \toprule
    Circuit & MATH & MATH-S & GSM-PLUS & SPELL-M & SPELL-MI & MATH-PARSE & LING-PARSE & LING-SV & LING-GA & LING-HJ \\
    \midrule
    MATH & 0.58 & 0.48 & 0.60 & 0.46 & 0.28 & 0.43 & 0.14 & 0.25 & -0.01 & 0.05 \\
    MATH-S & 0.52 & 0.51 & 0.51 & 0.42 & 0.25 & 0.45 & 0.30 & 0.30 & 0.02 & 0.06 \\
    GSM-PLUS & 0.61 & 0.49 & 0.92 & 0.88 & 0.71 & 0.43 & 0.50 & 0.28 & 0.05 & 0.08 \\
    SPELL-M & 0.41 & 0.42 & 0.69 & 0.98 & 0.91 & 0.44 & 0.64 & 0.39 & 0.04 & 0.21 \\
    SPELL-MI & 0.42 & 0.42 & 0.71 & 0.97 & 0.94 & 0.44 & 0.64 & 0.27 & 0.02 & 0.18 \\
    MATH-PARSE & 0.29 & 0.33 & 0.04 & 0.05 & 0.03 & 0.51 & 0.26 & 0.07 & 0.02 & 0.05 \\
    LING-PARSE & 0.34 & 0.30 & 0.18 & 0.40 & 0.25 & 0.12 & 0.75 & 0.16 & 0.05 & 0.15 \\
    LING-SV & 0.23 & 0.25 & 0.03 & 0.01 & 0.01 & 0.40 & 0.08 & 1.60 & 0.49 & 0.27 \\
    LING-GA & 0.17 & 0.30 & 0.09 & 0.04 & 0.03 & 0.41 & -0.00 & 0.94 & 0.99 & 0.58 \\
    LING-HJ & 0.34 & 0.31 & 0.17 & 0.04 & 0.03 & 0.41 & 0.00 & 0.56 & 0.20 & 0.79 \\
    \bottomrule
  \end{tabular}
  \caption{Llama-3.1-8B: Cross-task Faithfulness (normalized by clean/corrupted baselines; rows: circuits, columns: tasks; Top 2.5\%)}
  \label{table:llama-3.1-8b-faithfulness}
\end{table*}

\begin{table*}[t]
  \centering
  \scriptsize
  \tabcolsep=0.1cm
  \begin{tabular}{lrrrrrrrrrr}
    \toprule
    Circuit & MATH & MATH-S & GSM-PLUS & SPELL-M & SPELL-MI & MATH-PARSE & LING-PARSE & LING-SV & LING-GA & LING-HJ \\
    \midrule
    MATH & 0.61 & 0.49 & 0.86 & 0.12 & 0.13 & 0.44 & 0.25 & 0.03 & 0.00 & 0.09 \\
    MATH-S & 0.58 & 0.56 & 0.76 & 0.24 & 0.23 & 0.44 & 0.36 & 0.05 & 0.02 & 0.11 \\
    GSM-PLUS & 0.56 & 0.46 & 1.03 & 0.41 & 0.55 & 0.43 & 0.17 & 0.06 & 0.05 & 0.16 \\
    SPELL-M & 0.41 & 0.34 & 0.80 & 0.95 & 0.88 & 0.44 & 0.22 & 0.08 & 0.03 & 0.17 \\
    SPELL-MI & 0.48 & 0.40 & 0.84 & 0.80 & 0.93 & 0.43 & 0.33 & -0.00 & 0.02 & 0.12 \\
    MATH-PARSE & 0.12 & 0.12 & 0.10 & 0.03 & 0.04 & 0.46 & 0.04 & -0.05 & -0.00 & 0.05 \\
    LING-PARSE & 0.47 & 0.30 & 0.27 & 0.54 & 0.54 & 0.09 & 0.51 & 0.06 & 0.01 & 0.07 \\
    LING-SV & 0.28 & 0.22 & 0.23 & 0.05 & 0.03 & 0.42 & 0.08 & 1.04 & 0.60 & 0.15 \\
    LING-GA & 0.35 & 0.26 & 0.66 & 0.27 & 0.32 & 0.42 & 0.08 & 0.41 & 1.01 & 0.43 \\
    LING-HJ & 0.14 & 0.13 & 0.27 & 0.09 & 0.08 & 0.43 & 0.03 & 0.02 & 0.04 & 0.55 \\
    \bottomrule
  \end{tabular}
  \caption{Qwen3-1.7B: Cross-task Faithfulness (normalized by clean/corrupted baselines; rows: circuits, columns: tasks; Top 2.5\%)}
  \label{table:qwen3-1.7b-faithfulness}
\end{table*}

\begin{table*}[t]
  \centering
  \scriptsize
  \tabcolsep=0.1cm
  \begin{tabular}{lrrrrrrrrrr}
    \toprule
    Circuit & MATH & MATH-S & GSM-PLUS & SPELL-M & SPELL-MI & MATH-PARSE & LING-PARSE & LING-SV & LING-GA & LING-HJ \\
    \midrule
    MATH & 0.61 & 0.58 & 0.65 & 0.24 & 0.16 & 0.38 & 0.02 & -0.02 & 0.01 & 0.13 \\
    MATH-S & 0.61 & 0.60 & 0.71 & 0.39 & 0.35 & 0.37 & 0.11 & 0.01 & 0.02 & 0.16 \\
    GSM-PLUS & 0.54 & 0.52 & 0.95 & 0.66 & 0.71 & 0.36 & 0.18 & -0.02 & -0.01 & 0.22 \\
    SPELL-M & 0.52 & 0.49 & 0.57 & 0.98 & 0.94 & 0.37 & 0.31 & 0.02 & 0.05 & 0.19 \\
    SPELL-MI & 0.51 & 0.48 & 0.57 & 0.92 & 0.94 & 0.43 & 0.50 & 0.09 & 0.05 & 0.19 \\
    MATH-PARSE & 0.14 & 0.18 & -0.14 & 0.02 & 0.02 & 0.54 & 0.16 & -0.10 & 0.00 & 0.04 \\
    LING-PARSE & 0.22 & 0.25 & -0.02 & 0.34 & 0.40 & 0.09 & 0.68 & -0.04 & 0.03 & 0.15 \\
    LING-SV & 0.32 & 0.30 & 0.02 & 0.04 & 0.04 & 0.38 & -0.01 & 1.07 & 0.60 & 0.33 \\
    LING-GA & 0.19 & 0.19 & -0.03 & 0.05 & 0.09 & 0.38 & 0.08 & 0.09 & 1.01 & 0.41 \\
    LING-HJ & 0.26 & 0.28 & 0.11 & 0.10 & 0.08 & 0.37 & -0.02 & 0.01 & 0.10 & 0.74 \\
    \bottomrule
  \end{tabular}
  \caption{Qwen3-4B: Cross-task Faithfulness (normalized by clean/corrupted baselines; rows: circuits, columns: tasks; Top 2.5\%)}
  \label{table:qwen3-4b-faithfulness}
\end{table*}

\begin{table*}[t]
  \centering
  \scriptsize
  \tabcolsep=0.1cm
  \begin{tabular}{lrrrrrrrrrr}
    \toprule
    Circuit & MATH & MATH-S & GSM-PLUS & SPELL-M & SPELL-MI & MATH-PARSE & LING-PARSE & LING-SV & LING-GA & LING-HJ \\
    \midrule
    MATH & 0.60 & 0.53 & 0.80 & 0.40 & 0.40 & 0.36 & 0.04 & -0.01 & 0.01 & 0.07 \\
    MATH-S & 0.61 & 0.57 & 0.77 & 0.57 & 0.59 & 0.30 & 0.05 & -0.01 & 0.03 & 0.07 \\
    GSM-PLUS & 0.60 & 0.54 & 0.99 & 0.77 & 0.76 & 0.36 & 0.15 & 0.00 & 0.04 & 0.15 \\
    SPELL-M & 0.48 & 0.42 & 0.78 & 0.98 & 0.94 & 0.40 & 0.45 & 0.10 & 0.03 & 0.12 \\
    SPELL-MI & 0.54 & 0.48 & 0.82 & 0.97 & 0.99 & 0.47 & 0.28 & 0.03 & 0.05 & 0.15 \\
    MATH-PARSE & 0.24 & 0.24 & 0.14 & 0.05 & 0.06 & 0.57 & 0.14 & -0.06 & 0.00 & 0.06 \\
    LING-PARSE & 0.20 & 0.22 & 0.24 & 0.47 & 0.45 & 0.18 & 0.69 & -0.03 & 0.05 & 0.10 \\
    LING-SV & 0.24 & 0.23 & 0.23 & 0.06 & 0.04 & 0.31 & 0.03 & 1.13 & 0.48 & 0.25 \\
    LING-GA & 0.23 & 0.25 & 0.28 & 0.11 & 0.14 & 0.35 & 0.03 & 0.22 & 1.01 & 0.53 \\
    LING-HJ & 0.29 & 0.26 & 0.33 & 0.10 & 0.09 & 0.31 & -0.06 & 0.07 & 0.08 & 0.94 \\
    \bottomrule
  \end{tabular}
  \caption{Qwen3-8B: Cross-task Faithfulness (normalized by clean/corrupted baselines; rows: circuits, columns: tasks; Top 2.5\%)}
  \label{table:qwen3-8b-faithfulness}
\end{table*}

\begin{table*}[t]
  \centering
  \scriptsize
  \tabcolsep=0.1cm
  \begin{tabular}{lrrrrrrrrrr}
    \toprule
    Circuit & MATH & MATH-S & GSM-PLUS & SPELL-M & SPELL-MI & MATH-PARSE & LING-PARSE & LING-SV & LING-GA & LING-HJ \\
    \midrule
    MATH & 0.41 & 0.36 & 0.29 & 0.21 & 0.23 & 0.17 & 0.23 & -0.14 & 0.00 & 0.01 \\
    MATH-S & 0.26 & 0.31 & 0.04 & 0.22 & 0.28 & 0.17 & 0.23 & -0.16 & -0.00 & -0.00 \\
    GSM-PLUS & 0.31 & 0.29 & 0.62 & 0.34 & 0.36 & 0.22 & 0.34 & -0.13 & -0.00 & 0.02 \\
    SPELL-M & 0.25 & 0.22 & 0.26 & 0.92 & 0.97 & 0.17 & 0.82 & -0.05 & 0.01 & 0.06 \\
    SPELL-MI & 0.23 & 0.21 & 0.25 & 0.85 & 0.97 & 0.19 & 1.14 & 0.12 & 0.03 & 0.05 \\
    MATH-PARSE & 0.03 & 0.05 & -0.14 & 0.01 & 0.00 & 0.24 & 0.05 & -0.21 & 0.01 & 0.01 \\
    LING-PARSE & 0.01 & 0.01 & -0.12 & 0.14 & 0.20 & -0.04 & 1.26 & -0.07 & 0.01 & 0.02 \\
    LING-SV & 0.02 & 0.02 & -0.15 & 0.01 & 0.01 & 0.22 & -0.05 & 1.09 & 0.07 & 0.08 \\
    LING-GA & 0.07 & 0.09 & -0.12 & 0.01 & 0.00 & 0.21 & 0.02 & 0.30 & 0.88 & 0.09 \\
    LING-HJ & 0.09 & 0.08 & -0.00 & 0.08 & 0.05 & 0.26 & 0.11 & -0.07 & 0.05 & 0.43 \\
    \bottomrule
  \end{tabular}
  \caption{Gemma-2-2B: Cross-task Faithfulness (normalized by clean/corrupted baselines; rows: circuits, columns: tasks; Top 2.5\%)}
  \label{table:gemma-2-2b-faithfulness}
\end{table*}

\section{Sequence-Level Evaluation of Next-Token-Derived Circuits}
\label{sec:sequence-evaluation}
The main EAP-IG experiments extract and evaluate circuits using the first answer token.
For example, when the expected \textsc{Math-Parse} output is \texttt{2 * 5}, the logit-difference objective evaluates only its first token, \texttt{2}, rather than the complete subexpression.
To examine whether this restriction affects cross-task faithfulness, we use complete answer sequences to reevaluate the existing first-token EAP-IG circuits in Qwen3-8B for four tasks that can require multi-token outputs: \textsc{Math-s}, \textsc{Spell-m}, \textsc{Spell-mi}, and \textsc{Math-Parse}.
We do not extract circuits with a sequence-level attribution objective.
For input $x$, clean answer sequence $y$, and corrupted answer sequence $y'$, we compute the following metric:
\begin{multline}
m_{\mathrm{seq}}(x;y,y') =\\
\frac{1}{|y|}\sum_{t=1}^{|y|}\log p(y_t\mid x,y_{<t})\\
- \frac{1}{|y'|}\sum_{t=1}^{|y'|}\log p(y'_t\mid x,y'_{<t}).
\label{eq:sequence-metric}
\end{multline}

\begin{figure*}[t]
  \centering
  \includegraphics[width=0.72\textwidth]{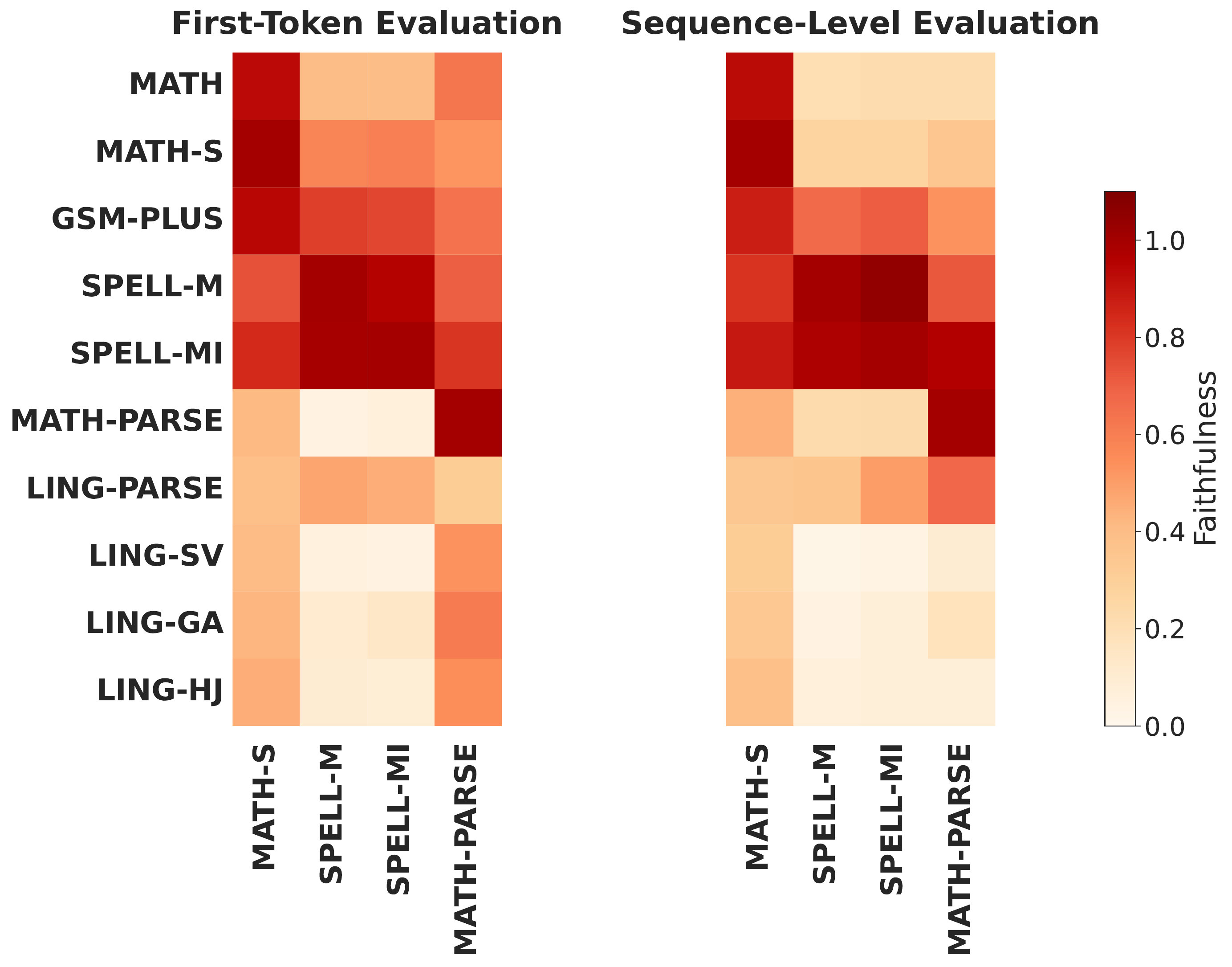}
  \caption{Comparison of first-token (left) and sequence-level (right) cross-task faithfulness for Qwen3-8B under the top-$2.5\%$ circuit definition. Rows denote the ten source-circuit tasks, and columns denote the four target tasks reevaluated at the sequence level. Both panels use the same circuits extracted and selected with first-token EAP-IG scores and share a color scale. Scores are normalized by clean/corrupted baselines and then by the corresponding target's self-circuit faithfulness within each evaluation setting.}
  \label{fig:sequence-faithfulness}
\end{figure*}

Figure~\ref{fig:sequence-faithfulness} compares first-token and sequence-level faithfulness for the four target tasks.
The transfer pattern for \textsc{Math-s} changes little.
\textsc{Spell-m} and \textsc{Spell-mi} also show broadly similar patterns, although faithfulness of the \textsc{Math-s} circuit is lower under sequence-level evaluation.
For \textsc{Math-Parse}, faithfulness decreases for most linguistic-task circuits, particularly \textsc{Ling-SV}, \textsc{Ling-GA}, and \textsc{Ling-HJ}; \textsc{Ling-Parse} is an exception.
This makes the separation between \textsc{Math-Parse} and these linguistic probes more pronounced.
One possible explanation is that scoring the complete first subexpression is more demanding than scoring its first token.

\section{Cross-Model Agreement of Hierarchical Clustering}
\label{sec:clustering-agreement}
\begin{figure*}[t]
  \centering
  \includegraphics[width=\textwidth]{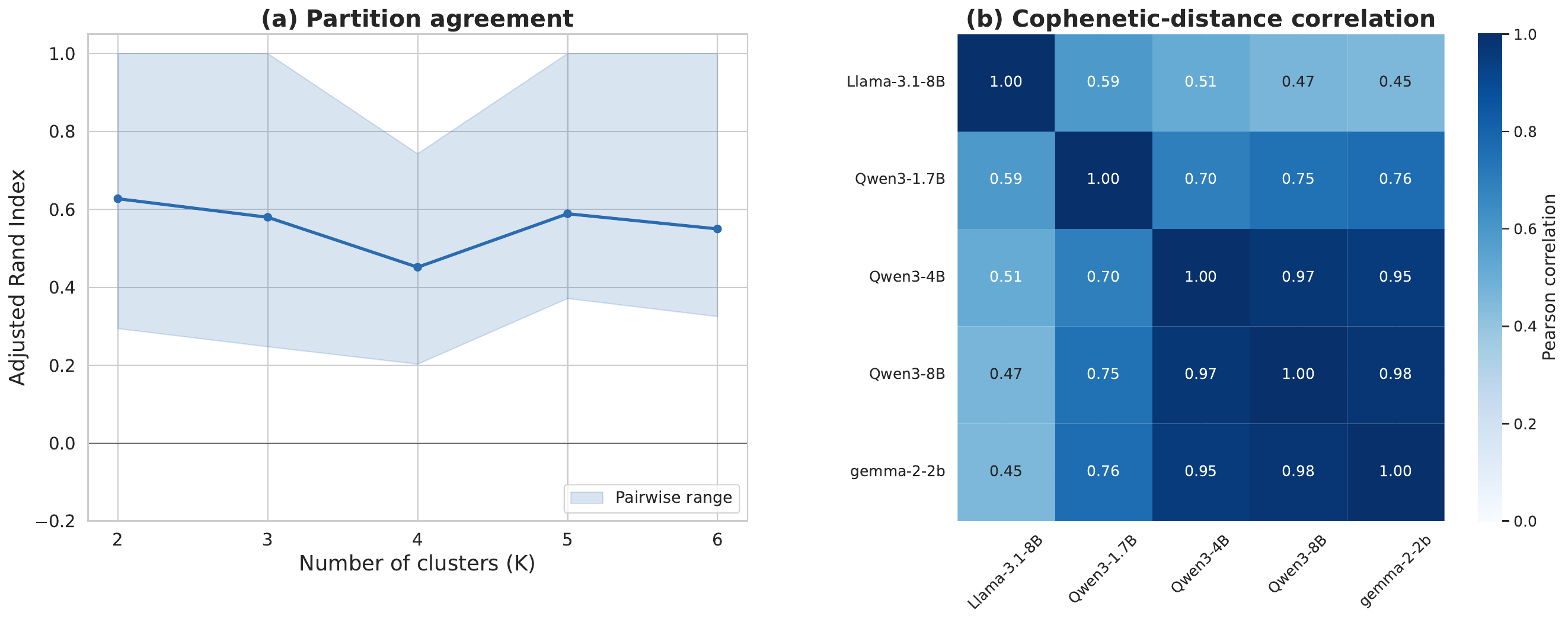}
  \caption{Cross-model agreement of hierarchical clustering based on source-circuit faithfulness profiles under the top-$2.5\%$ circuit definition. (a) Mean Adjusted Rand Index (ARI) over all ten model pairs after cutting each dendrogram into $K$ clusters; shading spans the pairwise minimum and maximum. (b) Pearson correlations between cophenetic-distance vectors, comparing complete dendrograms without selecting a cut level.}
  \label{fig:clustering-agreement}
\end{figure*}

Figure~\ref{fig:clustering-agreement} shows cross-model agreement of hierarchical clustering.
To compare full dendrograms without choosing a cut level for ARI, we also compute Pearson correlations between vectors of cophenetic distances over all task pairs.

Figure~\ref{fig:clustering-agreement} and Table~\ref{tab:clustering-ari} show moderate overall agreement across models, although their task partitions are not identical.
The cophenetic correlations are high among Qwen3-4B, Qwen3-8B, and Gemma-2-2B (0.95--0.98), whereas Llama-3.1-8B has lower correlations with every other model (0.45--0.59).
Qwen3-1.7B shows intermediate correlations with the other Qwen3 models and the similarly sized Gemma-2-2B (0.70--0.76).
The ARI results also show that partition agreement varies with the number of clusters.
Within these five models, clustering similarity does not show a simple relationship with either model family or parameter count.

\end{document}